\documentclass[10pt,twocolumn,letterpaper]{article}

\usepackage[pagenumbers]{cvpr} 

\usepackage[dvipsnames]{xcolor}

\usepackage{pifont}
\newcommand{\cmark}{\ding{51}}

\usepackage[accsupp]{axessibility} 
\usepackage{algorithm}
\usepackage{algpseudocode}
\usepackage{multicol}
\usepackage{multirow}
\usepackage{subcaption}
\usepackage{tikz}
\usetikzlibrary{bayesnet}

\definecolor{cvprblue}{rgb}{0.21,0.49,0.74}
\usepackage[pagebackref,breaklinks,colorlinks,citecolor=cvprblue]{hyperref}

\def\paperID{13146} 
\def\confName{CVPR}
\def\confYear{2024}

\title{Quantifying Uncertainty in Motion Prediction with Variational Bayesian Mixture}

\author{
    Juanwu Lu\thanks{Equal Contributions},\hspace{1em}
    Can Cui\footnotemark[1],\hspace{1em}
    Yunsheng Ma,\hspace{1em}
    Aniket Bera,\hspace{1em}
    Ziran Wang
    \\
    Purdue University, West Lafayette, USA\\
    {\tt\small \{juanwu, cancui, yunsheng, aniketbera, ziran\}@purdue.edu}
}

\begin{document}
\maketitle

\begin{abstract}
Safety and robustness are crucial factors in developing trustworthy autonomous vehicles. One essential aspect of addressing these factors is to equip vehicles with the capability to predict future trajectories for all moving objects in the surroundings and quantify prediction uncertainties. In this paper, we propose the Sequential Neural Variational Agent (SeNeVA), a generative model that describes the distribution of future trajectories for a single moving object. Our approach can distinguish Out-of-Distribution data while quantifying uncertainty and achieving competitive performance compared to state-of-the-art methods on the Argoverse 2 and INTERACTION datasets. Specifically, a 0.446 meters minimum Final Displacement Error, a 0.203 meters minimum Average Displacement Error, and a  5.35\% Miss Rate are achieved on the INTERACTION test set. Extensive qualitative and quantitative analysis is also provided to evaluate the proposed model. Our open-source code is available at \url{https://github.com/PurdueDigitalTwin/seneva}.   
\end{abstract}

\section{Introduction}
\label{sec: introduction}

Motion prediction is crucial for the safety and robustness of autonomous vehicles (AVs). The objective is to anticipate the potential future movements of the surrounding objects accurately. For the past few years, motion prediction has received emerging interests~\cite{liu_multimodal_2021,liang2020learning,janjos_bridging_2023,liao2022online,liu2022vision}, and we have seen significant progress in prediction accuracy on several benchmarks~\cite{zhan2019interaction,chang2019argoverse,sun2020scalability,wilson2023argoverse}. However, it remains a challenging task because the behaviors of traffic participants contain inherent multi-modal intentions and uncertainty. Therefore, instead of solely focusing on the prediction accuracy, it is also vital to identify modality and quantify the uncertainty about each predicted trajectory.

Existing methods account for the multi-modality primarily through generating multiple possible trajectories in parallel. Considering the procedure for generating the predictions, these methods mainly fall into two categories: sequential models and goal-based models. During inference, a sequential model directly forecasts a collection of possible future trajectories~\cite{liang2020learning, Gao_2020_CVPR}. Despite the prediction accuracy, most models fail to identify intentions and quantify uncertainty about predicted trajectories.

On the contrary, goal-based models generate a set of trajectory endpoint candidates, namely goals, and then complete the intermediate route connecting the object's current location to them~\cite{gilles_gohome_2022, densetnt,lu2024generalizable}. They share a common assumption that these sampled goals are multi-modal and account for most uncertainty. Although one can empirically extrapolate the intention behind each predicted goal, these models often rely on an expressive latent variable space that arbitrarily represents the mixture of all modalities without a specific architecture design that accounts for individual modality. Meanwhile, most also ignore quantifying prediction uncertainties or require recursive sampling to approximate uncertainty in trajectories~\cite{ABDAR2021243}, sharing a limitation similar to the direct-regression models.

This paper addresses these limitations by explicitly modeling the multi-modal trajectory distributions. Specifically, we propose a novel Bayesian mixture model, Sequential Neural Variational Agent (SeNeVA), that treats each observed trajectory in the dataset as being drawn from one of the generating processes. Each process has its own neural network parameterizing the distribution, and all the processes share a common upstream feature encoder. To improve the expressiveness of each process, we introduce a set of latent variables and analytically approximate their posteriors using variational inference. Distribution parameters directly quantify the prediction uncertainties, while the index of the generating process helps identify intention categories. In addition, we train a separate assignment network as a proxy to estimate the posterior distribution of mixture coefficients conditioned solely on the traffic condition. It helps promote generalization ability across different traffic scenarios. At inference time, we select one of the mixture components based on the probability determined by the assignment network and then sample trajectories specific to that particular mode in the scenario. Our experiment results show that the proposed model achieves competitive prediction accuracy against state-of-the-art methods with extensive information on intention and uncertainty associated with the predictions. The main contributions presented in this paper are: 
\begin{itemize}
    \item We address the limited uncertainty quantification in existing motion prediction models and propose a novel Bayesian mixture method to model the multi-modal distribution of future trajectories conditioned on historical traffic conditions.
    \item A separate assignment network and an NMS sampling method are introduced to allow for generating a small set of representative trajectories.
    \item An end-to-end training procedure is developed using variational inference, where the efficiency and robustness of our model are demonstrated through extensive ablations.
\end{itemize}

\begin{figure}[t]
    \centering
    \includegraphics[width=\linewidth]{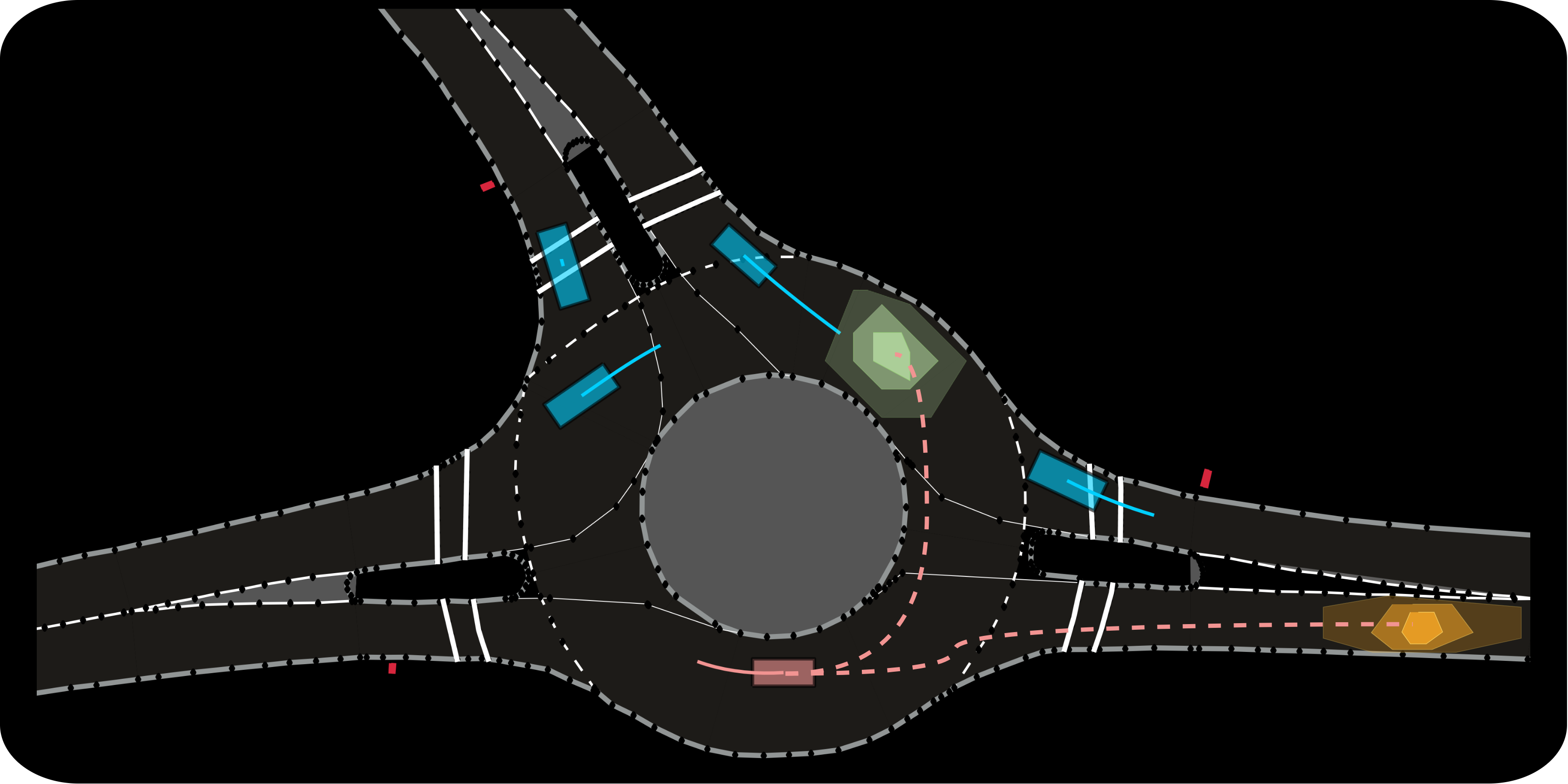}
    \caption{\textbf{Illustration of multi-modal trajectory distribution}. There are four surrounding vehicles (\textcolor{Blue}{blue}) and one target vehicle (\textcolor{Red}{red}) in the scene. The target vehicle can either turn left within the roundabout (\textcolor{LimeGreen}{green})  or turn right into the nearest exit (\textcolor{Orange}{orange}).}
    \label{fig: multi-modal}
\end{figure}
\section{Related Work}
\label{sec: related-work}

\subsection{Generative Motion Prediction Model}
\label{sec: generative-model}

Early works using deep neural networks for motion prediction commonly adopted recurrent networks~\cite{morton2016analysis,alahi2016social,vemula2018social} for time-series prediction. However, deterministic predictors have limited capability to capture multi-modal intentions. Recent advancements in generative models~\cite{kingma2022autoencoding,goodfellow2014generative,sohn2015learning} have demonstrated their promising power in producing diverse and realistic predictions, enabling a shift from learning a deterministic prediction model to fitting the distribution of all possible future trajectories. Some existing works apply Generative Adversarial Networks (GANs)~\cite{goodfellow2014generative} to fit a pair of generator-discriminator for trajectory prediction~\cite{gupta2018social,kosaraju2019social,sadeghian2019sophie,zhao2019multi}. Despite their promising performance, GANs can be unstable during training~\cite{kodali2017convergence} and lack interpretability of their generating processes. and lack interpretability in their generating processes. Other works adopt Gaussian Mixture Models (GMMs), leveraging the multi-modal property of mixture models~\cite{ivanovic2019trajectron,ivanovic2018generative,lee_desire_2017,pmlr-v100-chai20a}. Meanwhile, existing works explore using variational autoencoders (VAEs) for motion prediction~\cite{schmerling2018multimodal, ivanovic2018generative, salzmann2020trajectron++}. However, these works often directly use distribution means as predictions and neglect quantifying uncertainties in the predictions. Our approach addresses these limitations with a variational Bayes mixture model and investigates the prediction uncertainty quantified by its parameters.

\subsection{Uncertainty Quantification}

Uncertainty quantification has emerged as an area of interest since learning-based methods can be unreliable when data are out of the distribution of training samples. In autonomous driving, existing works have explored applying uncertainty quantification in object detection~\cite{feng2021review,hall2020probabilistic,chiu2021probabilistic} to improve perception robustness. Meanwhile, a few works investigate uncertainty quantification in trajectory prediction tasks. Djuric et al.~\cite{Djuric_2020_WACV} account for the inherent uncertainty of motions in traffic and use CNN in short-term motion prediction. Gaussian Process regression is also an alternative method in motion prediction tasks~\cite{4359316} as it can quantify uncertainty. Wang et al.~\cite{WANG2021107650} apply a Bayesian-entropy method considering uncertainty for predicting accurate trajectories and accidents. A comprehensive review of existing works can be found in~\cite{ABDAR2021243}. Nevertheless, uncertainty quantification in motion prediction remains underexplored.

\begin{figure*}[t!]
    \centering
    \includegraphics[width=\textwidth]{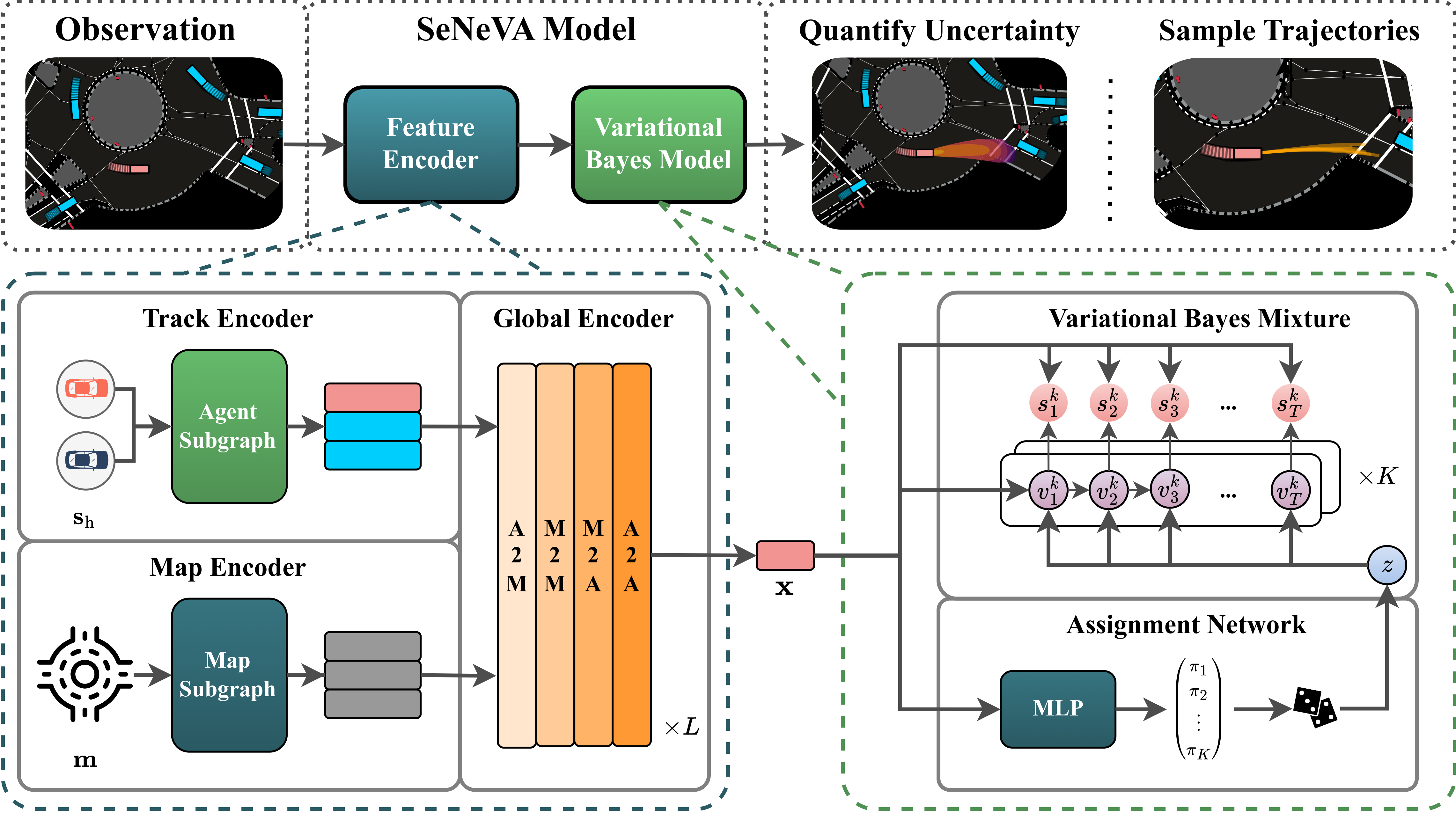}
    \caption{\textbf{Architecture of the proposed SeNeVA model}. The track and map encoders encode HD map and agent history trajectories. A global encoder module with a cascade of multi-head attention layers passes messages between map and agents to compute the context feature $\boldsymbol{x}$ from the perspective of the target agent. A variational Bayes Model of $K$ components estimates the distribution $p(\boldsymbol{s}_f|\boldsymbol{x})$ of trajectories conditioned on the context feature $x$. Additionally, we have an assignment network to estimate the distribution of mixture coefficients $p(z|\boldsymbol{x})$ conditioned on the context feature. The estimated distributions quantify the uncertainty of all possible future trajectories and enable the sampling of representative ones.}
    \label{fig: model}
\end{figure*}
\section{Problem Statement}
\label{sec: problem-statement}

This paper aims to address the problem of predicting the future trajectories of a single traffic participant in the scene and quantifying prediction uncertainties. The model derives predictions conditioned on the observation history. Suppose the observation horizon is $H$, and the prediction horizon is $T$. Given the history motion states $\boldsymbol{s}_\text{h}^{(i)}=\left[s_1^{(i)},s_2^{(i)},\ldots,s_H^{(i)}\right]$ of a single agent $i$, the objective is to predict its future motion states $\boldsymbol{s}^{(i)}_\text{f}=\left[s_{H+1}^{(i)},s_{H+2}^{(i)},\ldots,s_{H+T}^{(i)}\right]$, conditioned on the observation history of its surroundings $\boldsymbol{m}_\text{h}=\left[m_1,m_2,\ldots,m_H\right]$. To enable uncertainty quantification, instead of learning a model that outputs deterministic predictions, we build our model to estimate the conditional probability $p(\boldsymbol{s}^{(i)}_\text{f}|\boldsymbol{m}_\text{h},\boldsymbol{s}^{(i)}_\text{h})$.

However, the conditional distribution in real-world cases can be multi-modal. The modes can be spatially disjoint in some circumstances since they reflect mutually exclusive intentions. The example in Figure~\ref{fig: multi-modal} illustrates such a situation where a vehicle in the roundabout can either keep cruising within the roundabout or head towards the nearest exit. Existing models that generate predictions and uncertainties using features from a unified latent space can derive inaccurate trajectory predictions between any pair of spatially disjoint modes. Therefore, we decompose the conditional distribution as a mixture of $K$ disjoint components to promote accurate quantification of the uncertainties:

\begin{equation}
\label{eq: decomp-conditional}
p(\boldsymbol{s}^{(i)}_\text{f}|\boldsymbol{m}_\text{h},\boldsymbol{s}^{(i)}_\text{h})=\sum\limits_{k=1}^K p(\boldsymbol{s}^{(i)}_\text{f}|\boldsymbol{m}_\text{h},\boldsymbol{s}^{(i)}_\text{h},z=k)p(z=k),
\end{equation}
where $z$ is an indicator variable for possible futures. We will drop the superscripts $i$ in the following sections to simplify our notation. Following this formulation, we further make the below assumptions:

\begin{itemize}
\item The ground-truth trajectory in each training data is a sample from one of the mixture components reflecting its associated intention.
\item The residuals concerning the trajectory predictions follow a Multivariate normal distribution in space.
\item The displacements in space between consecutive time steps form a temporally correlated time series.
\end{itemize}

Our proposed model reflects the formulation and the assumptions with three key designs. First, we model each component trajectory distribution as a Multivariate Gaussian; that is, we quantify the uncertainty of each prediction as the residual. Then, the ground-truth trajectory follows a Bayes Mixture of individual Gaussian distributions with only one of the components activated for each observation. Finally, we introduce an extra latent variable to capture the temporal correlations of the time series.

Compared to other existing methods, our method directly models the entire distribution of all possible futures, providing rich information for downstream tasks such as risk and safety analysis. We incorporate variational inference to train our model. During model inference, the trained probabilistic model supports direct sampling on the mixture distribution using selection methods such as Non-Maximum Suppression~\cite{neubeck2006efficient,hosang2017learning} for applications requiring only a small set of representative trajectories.
\section{Sequential Neural Variational Agent}
\label{sec: method}

This section presents the Sequential Neural Variational Agent (SeNeVA) model for single-agent motion prediction (see Figure~\ref{fig: model}). Our model learns to use encoded features of the traffic environment (Section~\ref{sec: feature-encoding}) to parameterize a spatial distribution of plausible trajectories as a mixture of Multivariate Gaussian distributions (Section~\ref{sec: variational-bayes}). In addition, we implement an assignment network to estimate the mixture weights (Section~\ref{sec: z-proxy}), which aims to avoid repeated sampling of latent variables and improve the generalization performance in unseen cases. Finally, we introduce the training objective (Section~\ref{sec: training}) and how we sample from the distribution (Section~\ref{sec: sampling}).

\subsection{Feature Encoding}
\label{sec: feature-encoding}

We represent the history of the surrounding environment and agent motions using a vectorized representation. Specifically, the history motion states of agent $i$ are represented by a vector $\boldsymbol{s}_h^{(i)}\in\mathbb{R}^{H\times 5}$ consisting of the locations, heading, and velocities at each time step. We consider the surrounding to be a static HD map represented by a collection of $p$ polylines $\boldsymbol{m}_\text{h}=\left\{l_1,\ldots,l_p\right\}$, where each polyline is a set of vectors $l_p\in\mathbb{R}^{N_p\times 4}$, each denoted by the coordinates of its head and tail. All coordinates are projected into a target-centric frame.

We encode the map and agent history using two separate VectorNet subgraphs~\cite{Gao_2020_CVPR}, resulting in a polyline feature $\boldsymbol{p}_i$ for each agent and each polyline on the map. To model high-level interaction, we follow LaneGCN~\cite{liang2020learning} and model four types of global interactions, including agent-to-map-polyline (A2M), map-polyline-to-map-polyline (M2M), map-polyline-to-agent (M2A), and agent-to-agent (A2A) interaction. We use four individual multi-head attention (MHA) layers at each global interaction level to model each type of interaction individually and use a cascade of $L$-level MHA layers to encode global interactions. The implementation of the encoder module is provided in the supplementary material.

As a result, the output $\boldsymbol{x}$ from the encoder module is a latent representation of the traffic condition from the perspective of the target agent. The consecutive probabilistic model learns the conditional distribution $p(\boldsymbol{s}_\text{f}|\boldsymbol{x},z)$ as the equivalence for $p(\boldsymbol{s}_\text{f}|\boldsymbol{m}_\text{h},\boldsymbol{s}_\text{h},z)$ in equation~\ref{eq: decomp-conditional}.
\subsection{Variational Bayes Mixture}
\label{sec: variational-bayes}

Instead of directly predicting a sequence of future locations, our model predicts the displacements between consecutive time steps for stability. We model these displacements as a time series. To capture the temporal dependencies, we introduce a latent variable $\boldsymbol{v}=\left[v_{H+1},v_{H+2},\ldots,v_{H+T}\right]$ and factorize the conditional distribution as $p(\boldsymbol{s}_\text{f}|\boldsymbol{x},z)=\int_{\boldsymbol{v},\boldsymbol{x}}p(\boldsymbol{s}_\text{f}|\boldsymbol{v},\boldsymbol{x})\cdot p(\boldsymbol{v}|\boldsymbol{x},z)d{\boldsymbol{v}}$, where
\begin{equation}
\label{eq: psv}
p(\boldsymbol{s}_\text{f}|\boldsymbol{v},\boldsymbol{x})=\prod\limits_{t=1}^{T}p(s_{H+t}|v_{H+t},\boldsymbol{x}),
\end{equation}
\begin{equation}
\label{eq: pvxz}
p(\boldsymbol{v}|\boldsymbol{x},z)=p(v_{H+1}|\boldsymbol{x},z)\prod\limits_{t=1}^{T-1}p(v_{H+t+1}|v_{H+t},\boldsymbol{x}).
\end{equation}

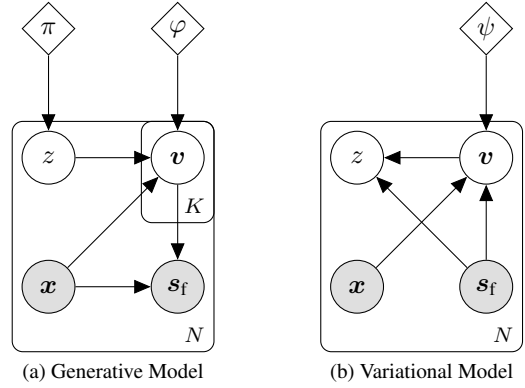
\begin{figure}[t]
    \centering
    \begin{subfigure}{.4\columnwidth}
        \centering
        \begin{tikzpicture}
            \node[obs] (s) {$\boldsymbol{s}_\text{f}$}; %
            \node[obs, left=of s] (x) {$\boldsymbol{x}$};
            \node[latent, above=of s] (v) {$\boldsymbol{v}$}; %
            \node[latent, left=of v] (z) {$z$}; %
            \node[det, above=of z] (pi) {$\pi$}; %
            \node[det, above=of v] (phi) {$\varphi$}; %
            \plate {} {(s)(x)(v)(z)} {$N$}; %
            \plate {} {(v)} {$K$}; %
            \edge {pi} {z};
            \edge {x} {v, s};
            \edge {v} {s};
            \edge {z} {v};
            \edge {phi} {v};
        \end{tikzpicture}
        \caption{Generative Model}
        \label{fig:likelihood-family}
    \end{subfigure}
    \hspace{.5em}
    \begin{subfigure}{.5\linewidth}
        \centering
        \begin{tikzpicture}
            \node[obs] (s) {$\boldsymbol{s}_\text{f}$}; %
            \node[obs, left=of s] (x) {$\boldsymbol{x}$};
            \node[latent, above=of s] (v) {$\boldsymbol{v}$}; %
            \node[latent, left=of v] (z) {$z$}; %
            \node[det, above=of v] (psi) {$\psi$}; %
            \plate {} {(s)(x)(v)(z)} {$N$}; %
            \edge {x} {v};
            \edge {s} {v, z};
            \edge {v} {z};
            \edge {psi} {v};
        \end{tikzpicture}
        \caption{Variational Model}
        \label{fig: variational-family}
    \end{subfigure}
    \caption{The graphical representation of the (a) generative model and the (b) variational family. Shaded and unshaded nodes are the observed and latent random variables. Diamond nodes are the model parameters.}
    \label{fig: graphical-model}
\end{figure}

As illustrated by Figure~\ref{fig: graphical-model}, we parameterize the generative process as a conditional variational model given by:
\begin{equation}
\label{eq: generative-model}
p_{\theta,\varphi}(\boldsymbol{s}_\text{f},\boldsymbol{v},z|\boldsymbol{x})=p_\theta(\boldsymbol{s}_\text{f}|\boldsymbol{v},\boldsymbol{x})\cdot p_\varphi(\boldsymbol{v}|\boldsymbol{x},z)\cdot p(z),
\end{equation}
where we consider the latent variables to follow the below processes:
\begin{subequations}
\begin{equation}
\label{eq: z-dist}
z\sim\text{Categorical}(\pi),
\end{equation}
\begin{equation}
\label{eq: v1-dist}
v_{H+1}|\boldsymbol{x},z\sim\prod\limits_{k=1}^K\mathcal{N}\left(\mu(\boldsymbol{x};\varphi_{0,k}),\text{diag}(\sigma^2(\boldsymbol{x};\varphi_{0,k}))\right)^{z_k},
\end{equation}
\begin{equation}
    v_{t+1}|v_t,\boldsymbol{x},z\sim\prod\limits_{k=1}^K\mathcal{N}\left(\mu(\boldsymbol{x};\varphi_{r,k}),\text{diag}(\sigma^2(\boldsymbol{x};\varphi_{r,k}))\right)^{z_k},
\end{equation}
\begin{equation}
\label{eq: s-dist}
s_t|v_t,\boldsymbol{x}\sim\mathcal{N}\left(\mu\left(\left[v_t,\boldsymbol{x}\right];\theta\right),\Sigma\left(\left[v_t,\boldsymbol{x}\right];\theta\right)\right).
\end{equation}
\end{subequations}
The $\mu(\cdot;\theta)$,$\Sigma(\cdot;\theta)$,$\mu(\cdot;\varphi_*)$, and $\Sigma(\cdot;\varphi_*)$ above are outputs from neural networks with parameters $\theta$ and $\varphi_*$. The $\left[\cdot,\cdot\right]$ denotes concatenation operation. The $\text{diag}\left[\cdot\right]$ is the operation to create a diagonal matrix with the given values. In practice, $\theta$ and $\varphi_{0,k}$ are parameters of a Multi-layer Perceptron (MLP), and $\varphi_{r,k}$ are parameters of a Long Short-Term Memory (LSTM)~\cite{memory2010long}. We use a non-informative, uniform prior $\pi_k=1/K, k=1, \ldots, K$ for the random variable $z$ since we assume the dataset covers all possible intentions equally. One of the key advantages of our formulation is that the use of LSTM cells and the MLP in equation~\ref{eq: s-dist} promotes parameter efficiency and enables the handling of variable-length trajectory predictions

However, fitting the generative model by directly maximizing the log-likelihood $\log p(\boldsymbol{s}_\text{h}|\boldsymbol{x})$ is intractable. We approach this problem using variational inference. Specifically, we use the mean-field variational family to approximate the true posterior, given as:
\begin{equation}
\label{eq: variational-family}
q(\boldsymbol{v},z|\boldsymbol{s}_\text{f},\boldsymbol{x})= q_\psi(\boldsymbol{v}|\boldsymbol{s}_\text{f},\boldsymbol{x})\cdot q_\varphi(z|\boldsymbol{v},\boldsymbol{x}),
\end{equation}
Similar to the generative model, we factorize the sequential dependency of $q_\psi(\boldsymbol{v}|\boldsymbol{x},\boldsymbol{s}_\text{f})$ and parameterize it using a combination of MLP and LSTM. The factorization structure of $q_\psi(\boldsymbol{v}|\boldsymbol{x},\boldsymbol{s}_\text{f})$ resembles the one in equation~\ref{eq: pvxz}, given as:
\begin{equation}
\label{eq: qvxs}
q_\psi(\boldsymbol{v}|\boldsymbol{x},\boldsymbol{s}_f)=q(v_{H+1}|\boldsymbol{x},\boldsymbol{s}_\text{f})\prod\limits_{t=1}^{T-1}q(v_{H+t+1}|v_{H+t},\boldsymbol{x},\boldsymbol{s}_\text{f})
\end{equation}

Meanwhile, we reparameterize the posterior for the random variable $z$ by:
\begin{equation}
\label{eq: z-posterior}
q_\varphi(z=j|\boldsymbol{v},\boldsymbol{x})=\frac{p(z=j)p_\varphi(\boldsymbol{v}|\boldsymbol{x},z=j)}{\sum\limits_{k=1}^Kp(z=k)p_\varphi(\boldsymbol{v}|\boldsymbol{x},z=k)}
\end{equation}
This avoids an additional $q(z)$ network while allowing gradients to backpropagate onto the parameter set $\varphi$. The derivation is in the supplementary. Nevertheless, the trade-off is that we need further Monte-Carlo sampling during inference to estimate the $z$-posterior. The following section introduces our solution to this issue with a proxy network to directly output $p(z|x)$.
\subsection{Assignment Network}
\label{sec: z-proxy}

Sampling from the latent space to estimate the posterior assignment $q(z|\boldsymbol{v}^{(i)},\boldsymbol{x})$ can be computationally intensive. We introduce an assignment network that approximates $p(z|x)$ conditioned solely on the input feature $x$ to address this. The network is parameterized as an MLP that outputs the log-likelihood of mixture components:

\begin{equation}
\log\hat{\pi} = \text{MLP}(\boldsymbol{x}).
\end{equation}

The estimated $\hat{\pi}$ can be used as the mixture coefficient for uncertainty quantification during inference. For evaluations based on a small set of trajectories, the assignment network can help identify the components most likely to be sampled instead of uniformly sampling from all the mixture components. This approach can significantly reduce computational complexity while maintaining the quality of the generated trajectories.
\subsection{Model Training}
\label{sec: training}

We train our variational Bayes model to maximize the Evidence Lower Bound (ELBO). Using the factorizations given by equation~\ref{eq: generative-model} and equation~\ref{eq: variational-family}, the ELBO objective can be written as:

\begin{equation}
\label{eq: elbo-expr}
\begin{aligned}
\mathcal{L}_\text{ELBO} &=\mathbb{E}_{q_\psi(\boldsymbol{v}|\boldsymbol{s}_\text{f},\boldsymbol{x})}\log p_\theta(\boldsymbol{s}_\text{f}|\boldsymbol{v},\boldsymbol{x}) \\
&-\mathbb{E}_{q_\varphi(\boldsymbol{z}|\boldsymbol{v},\boldsymbol{x})}D_\text{KL}\left(q_\psi(\boldsymbol{v}|\boldsymbol{s}_\text{f},\boldsymbol{x})\|p_\varphi(\boldsymbol{v}|\boldsymbol{x},z)\right) \\
&-\mathbb{E}_{q_\psi(\boldsymbol{v}|\boldsymbol{s}_\text{f},\boldsymbol{x})}D_\text{KL}\left(q_\varphi(z|\boldsymbol{v},\boldsymbol{x})\|p(z)\right),
\end{aligned}
\end{equation}
where $D_\text{KL}(\cdot\|\cdot)$ denotes the Kullback-Leibler divergence~\cite{10.1214/aoms/1177729694}. We apply Monte-Carlo sampling to estimate the expectations in the ELBO. Since we reparameterize the $z$-posterior in equation~\ref{eq: z-posterior}, we can directly compute the second term in the ELBO using the Monte-Carlo samples. For each sample $j$, we define $w_{jk}=q_\varphi(z=k|\boldsymbol{v}^{(j)},\boldsymbol{x})$ and $d_{jk}=D_\text{KL}\left(q_\psi(\boldsymbol{v}^{(j)}|\boldsymbol{s}_\text{f},\boldsymbol{x})\|p_\varphi(\boldsymbol{v}^{(j)}|\boldsymbol{x},z=k)\right)$, then

\begin{equation}
\label{eq: monte-carlo-kl-z}
\mathbb{E}_{q(\boldsymbol{z}|\boldsymbol{v},\boldsymbol{x})}D_\text{KL}\left(q(\boldsymbol{v}|\boldsymbol{s}_\text{f},\boldsymbol{x})\|p(\boldsymbol{v}|\boldsymbol{x},z)\right)\approx
\frac{1}{N_\text{mc}}\sum\limits_{j=1}^{N_\text{mc}}\sum\limits_{k=1}^Kw_{jk}\cdot d_{jk},
\end{equation}
where $N_\text{mc}$ is the number of Monte-Carlo samples. A detailed derivation of the ELBO is provided in the supplementary material.

To train the assignment network, we first marginalize the ground-truth assignment weights of the mixture components given the motion states $\boldsymbol{s}_f$ by:

\begin{equation}
\label{eq: psx-margin}
p(\boldsymbol{s}_\text{f}|\boldsymbol{x},z)=\int_{\boldsymbol{v}\sim q(\boldsymbol{v}|\boldsymbol{s}_f,\boldsymbol{x})}p(\boldsymbol{s}_f|\boldsymbol{v},\boldsymbol{x})p(\boldsymbol{v}|\boldsymbol{x},z)d\boldsymbol{v}.
\end{equation}

We approximate the marginalization by applying Monte-Carlo sampling on the posterior distribution $q(\boldsymbol{v}|\boldsymbol{s}_f,\boldsymbol{x})$. Since we are using a non-informative uniform prior distribution for $z$, we can obtain the target assignment weights by applying Bayes' rule:

\begin{equation}
\label{eq: z-proxy-target}
p(z=j|\boldsymbol{x})=\frac{p(\boldsymbol{s}_f|\boldsymbol{x},z=j)p(z=j)}{p(\boldsymbol{s}_\text{f}|\boldsymbol{x})}=\frac{p(\boldsymbol{s}_\text{f}|\boldsymbol{x},z=j)}{\sum\limits_{k=1}^Kp(\boldsymbol{s}_\text{f}|\boldsymbol{x},z=k)}.
\end{equation}

We train the assignment network to minimize the focal loss~\cite{Lin_2017_ICCV} given by:

\begin{equation}
\label{eq: focal-loss}
\mathcal{L}_{\hat{\pi}} = -\sum\limits_{k=1}^K(1-\hat{\pi}_k)^\gamma\cdot p(z=k|\boldsymbol{x})\log\hat{\pi}_k,
\end{equation}
where $\gamma$ is a tunable focusing hyperparameter balancing well-classified and misclassified components. Overall, we train the SeNeVA model to minimize the weighted sum of two losses:

\begin{equation}
\mathcal{L} = -\mathcal{L}_\text{ELBO} + \alpha\mathcal{L}_{\hat{\pi}}.
\end{equation}
\subsection{Trajectory Sampling}
\label{sec: sampling}

The output from the SeNeVA model is the distribution of all possible trajectories in the future. However, many existing motion prediction challenges and applications require only a small set of the most probable predictions for evaluation. To promote the leverage of the distribution information, we propose a method to sample from the generative model with Non-Maximum Suppression (NMS). Denote $y_t=\sum\limits_{t^\prime=1}^ts_t$ as the location of the target agent at time $t$. Since $s_t$ are Gaussian random variables, the sum of them also follows a Gaussian distribution:

\begin{equation}
\label{eq: yt-dist}
y_t|v_{\leq t},\boldsymbol{x},z\sim\mathcal{N}\left(\sum\limits_{t^\prime=1}^t\mu([v_t ,\boldsymbol{x}];\theta),\sum\limits_{t^\prime=1}^t\Sigma([v_t,\boldsymbol{x}];\theta)\right).
\end{equation}

During sampling, we obtain the mixture weights approximation $\hat{\pi}$ from the assignment network to help determine which component we should sample from. Since the final location $y_{H+T}$ holds the most uncertainty, quantified by its covariance matrix as a summation of displacement covariance over all previous time steps, we propose first sampling $y_{H+T}$ using NMS, as described in Algorithm~\ref{alg: sampling}. For each sampled $y_{H+T}$, we generate the intermediate path from the target agent's current location $y_H$ to the sampled final destination $y_{H+T}$ by assuming uniform uncertainty over time. This assumption helps guarantee the smoothness of the sampled trajectories. Details are provided in the supplementary.

\begin{algorithm}[t]
\caption{Destination Sampling}
\label{alg: sampling}
\begin{algorithmic}[1]
\Require List of candidates $\{y_{H+T}\}$, Candidate buffer radius $r$, Intersection-over-Union (IoU) Threshold $\gamma$, Number of trajectories to sample $M$
\State Evaluate probabilities $p(y_{H+T}|v_{\leq T},\boldsymbol{x},z)$
\State Sort candidates by probabilities in a descending order
\State Initialize an empty list $Q$
\While{$\text{size}(Q)<M$}
    \State Take the most probable candidate $y_{H+T}^*$
    \State Add $y_{H+T}^*$ to $Q$
    \State Create circle $c$ centered at $y_T^*$ of radius $r$
    \For{each other candidate $y_{H+T}^\prime$}
        \State Create circle $c^\prime$ centered at $y_{H+T}^\prime$ of radius $r$
        \If{IoU$(c, c^\prime) > \gamma$}
            \State Remove $y_{H+T}^\prime$ from the list of candidates
        \EndIf
    \EndFor
    \State Remove $y_{H+T}^*$ from the list of candidates
\EndWhile
\State \textbf{return} $Q$
\end{algorithmic}
\end{algorithm}

\begin{table*}[t!]
    \centering
    \caption{Comparison of the proposed SeNeVA with various state-of-the-art methods on the two datasets. The top-performing method for each setting is highlighted in \textbf{bold}. The second-top-performing method is highlighted with an underscore (\_).}
    \begin{tabular}{@{}l|c|c|ccccccc@{}}
        \toprule
        \multicolumn{1}{c}{Dataset} \vrule & Split & Method & \# Param. & $\text{minFDE}_6$ $(\downarrow)$ & $\text{minADE}_6$ $(\downarrow)$& MR $(\downarrow)$\\
        \midrule
        \multirow{6}{*}{INTERACTION} 
        &\multirow{6}{*}{test} & DenseTNT~\cite{densetnt} &- & 0.795 & 0.424 & 0.060 \\
         & & Multi-Branch SS-ASP~\cite{janjos_bridging_2023} & - & 0.539 & \underline{0.178} & 0.115 \\
         & &MultiModalTransformer~\cite{huang_multi-modal_2022} & 6.3M & \text{0.551}& 0.213 & \textbf{0.051} \\
        & & HDGT~\cite{jia_hdgt_2022} & 15.3M& \underline{0.478} & \textbf{0.168} & 0.056 \\
         & & \textbf{SeNeVA (ours)} & \textbf{1.3M} & \textbf{0.446} & 0.203 & \underline{0.053} \\
        \midrule
        \multirow{4}{*}{INTERACTION} &\multirow{4}{*}{val}
        & DESIRE~\cite{lee_desire_2017} & - & 0.880 & 0.320 & -\\
        & & MultiPath~\cite{pmlr-v100-chai20a} & - & 0.990 &0.300 & -\\
        & & TNT~\cite{zhao_tnt_nodate} & - & \underline{0.670} & \underline{0.210} & -\\
        &  & \textbf{SeNeVA (ours)} & \textbf{1.3M} & \textbf{0.431} &\textbf{0.197} & \textbf{0.079}\\
        \midrule
        \multirow{3}{*}{Argoverse 2} & \multirow{3}{*}{val} 
        & DenseTNT~\cite{densetnt} &- & 1.620 & 0.960 & 0.233 \\
        & & Forecast-MAE~\cite{cheng_forecast-mae_nodate} & 1.9M & \underline{1.409} & \underline{0.901} & \underline{0.178} \\
        & & \textbf{SeNeVA (ours)} & \textbf{1.3M}  &  \textbf{1.319}&  \textbf{0.713}& \textbf{0.175} \\
        \bottomrule
    \end{tabular}
\label{tab:sota}
\end{table*}

\section{Experiments}
\label{sec: experiment}


\subsection{Experiment Setup}
\label{sec: setup}

\paragraph{Datasets} The evaluation is conducted on two benchmark datasets: the INTERACTION~\cite{zhan2019interaction} and the Argoverse 2~\cite{wilson2023argoverse} dataset. The INTERACTION dataset consists of data collected from $18$ different locations globally. The goal is to predict the 3-second future conditioned on a 1-second history observation. There are about $35\%$ Out-of-Distribution (OOD) data in the test dataset (i.e., with unseen map and traffic conditions). This is an interesting feature we leverage to analyze the uncertainty quantification performance of our proposed SeNeVA model. The Argoverse 2 dataset contains $250,000$ scenarios collected from $6$ different cities. The task is to predict a future 6-second trajectory based on a 5-second history observation.

\paragraph{Metrics} For uncertainty quantification, we evaluate the OOD identification capability of SeNeVA by comparing the predicted distribution entropy under in-distribution and OOD cases. For predicting a small set of representative trajectories, we use standard evaluation metrics, including minimum Average Displacement Error (minADE$_k$), minimum Final Displacement Error (minFDE$_k$), and Miss Rate (MR$_k$). We present further details about the metrics in the supplementary material.

\subsection{Uncertainty Quantification}
\label{sec: uncertainty}

\subsubsection{Quantitative Analysis}
\label{sec: uncertainty-quantitative}

The output trajectories from the SeNeVA model follow a bivariate Gaussian distribution, allowing the total uncertainty about the prediction to be directly measured by its entropy (calculation details in supplementary). We expect the SeNeVA model to have lower entropy for in-distribution cases and higher entropy for out-of-distribution (OOD) ones.

\begin{figure}[t!]
\centering
\includegraphics[width=\linewidth]{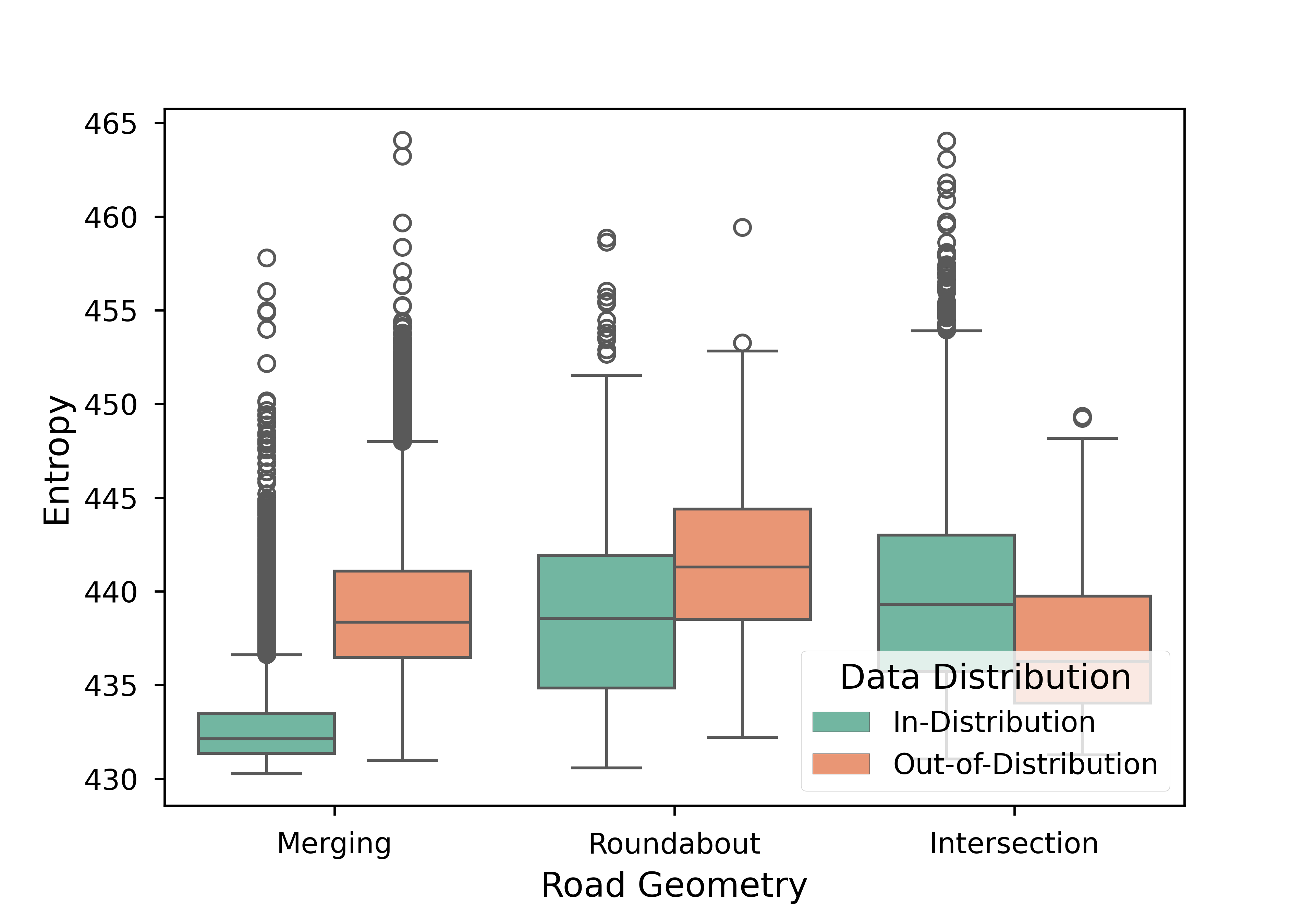}
\caption{\textbf{Predicted uncertainty in different road geometry and data distributions.} The predicted uncertainty in OOD cases is generally higher than in in-distribution cases.}
\label{fig: entropy}
\end{figure}

We compute the distribution entropy for the $22,498$ cases of the INTERACTION test dataset, where $7,898$ data cases are from unseen locations. Figure~\ref{fig: entropy} shows that the predicted total uncertainty in OOD cases is generally higher than in in-distribution data, with a $1.5\%$ increase in highway merging cases and $0.6\%$ increase in roundabout cases. This demonstrates SeNeVA's ability to distinguish OOD data during prediction and assign higher uncertainty to these cases. However, the OOD entropy in intersection cases is $0.5\%$ lower than in-distribution cases, possibly due to the more complex traffic conditions in intersections, including cyclists and pedestrians.

We further compare our model with the Deep Ensembles method~\cite{NIPS2017_9ef2ed4b} for uncertainty quantification, using VectorNet as the base ensembled model for fair comparison. Table \ref{tab: uq-compare} shows that SeNeVA can better distinguish OOD data, with an average inference time of $51$ms compared to $1,590$ms for the Deep Ensembles model, indicating better efficiency.
\begin{table}[t]
    \centering
    \caption{Predictive uncertainty measured by total entropy in in-distribution (ID) and out-of-distribution (OOD) cases.}
    \resizebox{0.95\linewidth}{!}{
    \begin{tabular}{cccc}
    \hline
    \multirow{2}{*}{\textbf{Model}}     & \multirow{2}{*}{\textbf{Geometry}} & \multicolumn{2}{c}{\textbf{Entropy}}                    \\ \cline{3-4} 
                                        &                                         & \textbf{ID} & \textbf{OOD} \\ \hline
    \multirow{3}{*}{\begin{tabular}[c]{@{}c@{}}VectorNet\\ Ensembles\end{tabular}} & Merging                                 & 90.98 (5.26)             & 87.69 (8.97)                \\
                                        & Roundabout                              & 95.59 (7.63)             & 104.62 (8.39)                \\
                                        & Intersection                            & 92.73 (6.81)             & 76.37 (5.35)                 \\
    \hline
    \multirow{3}{*}{\begin{tabular}[c]{@{}c@{}}SeNeVA\\ (Ours)\end{tabular}}      & Merging                                 & 422.64 (2.64)            & 429.28 (4.32)                \\
                                        & Roundabout                              & 428.12 (4.63)            & 431.17 (4.47)                \\
                                        & Intersection                            & 429.60 (5.63)            & 427.13 (4.33)                \\ \hline
    \end{tabular}
    }
    \label{tab: uq-compare}
\end{table}

\subsubsection{Qualitative Analysis}
\label{sec: uncertainty-qualitative}

\begin{figure*}[t!]
\centering
\begin{subfigure}{0.4\textwidth}
    \centering
    \includegraphics[width=\textwidth]{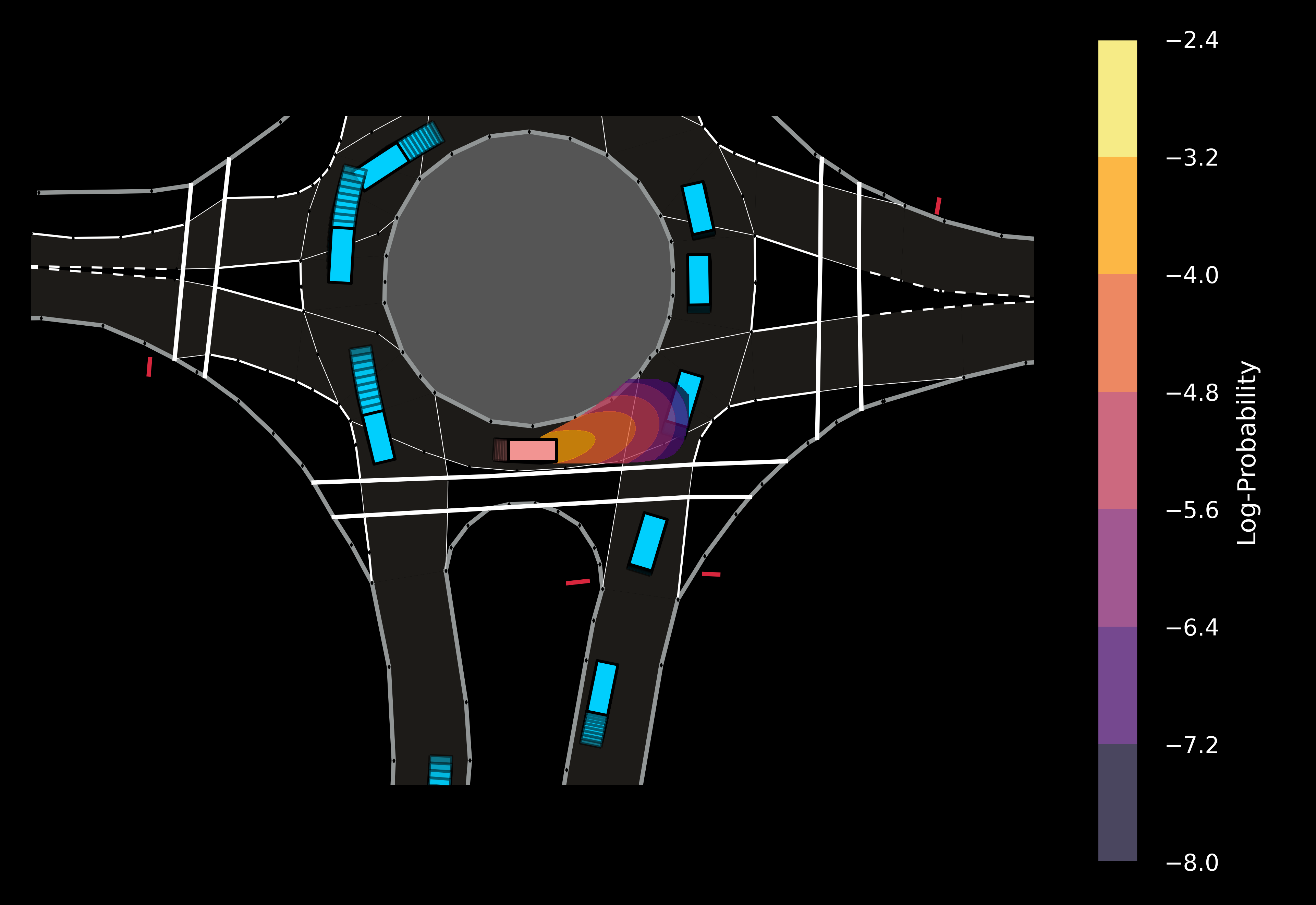}
    \caption{In-Distribution Case (DR\_USA\_Roundabout\_SR)}
\end{subfigure}
\hspace{1em}
\begin{subfigure}{0.4\textwidth}
    \centering
    \includegraphics[width=0.92\textwidth]{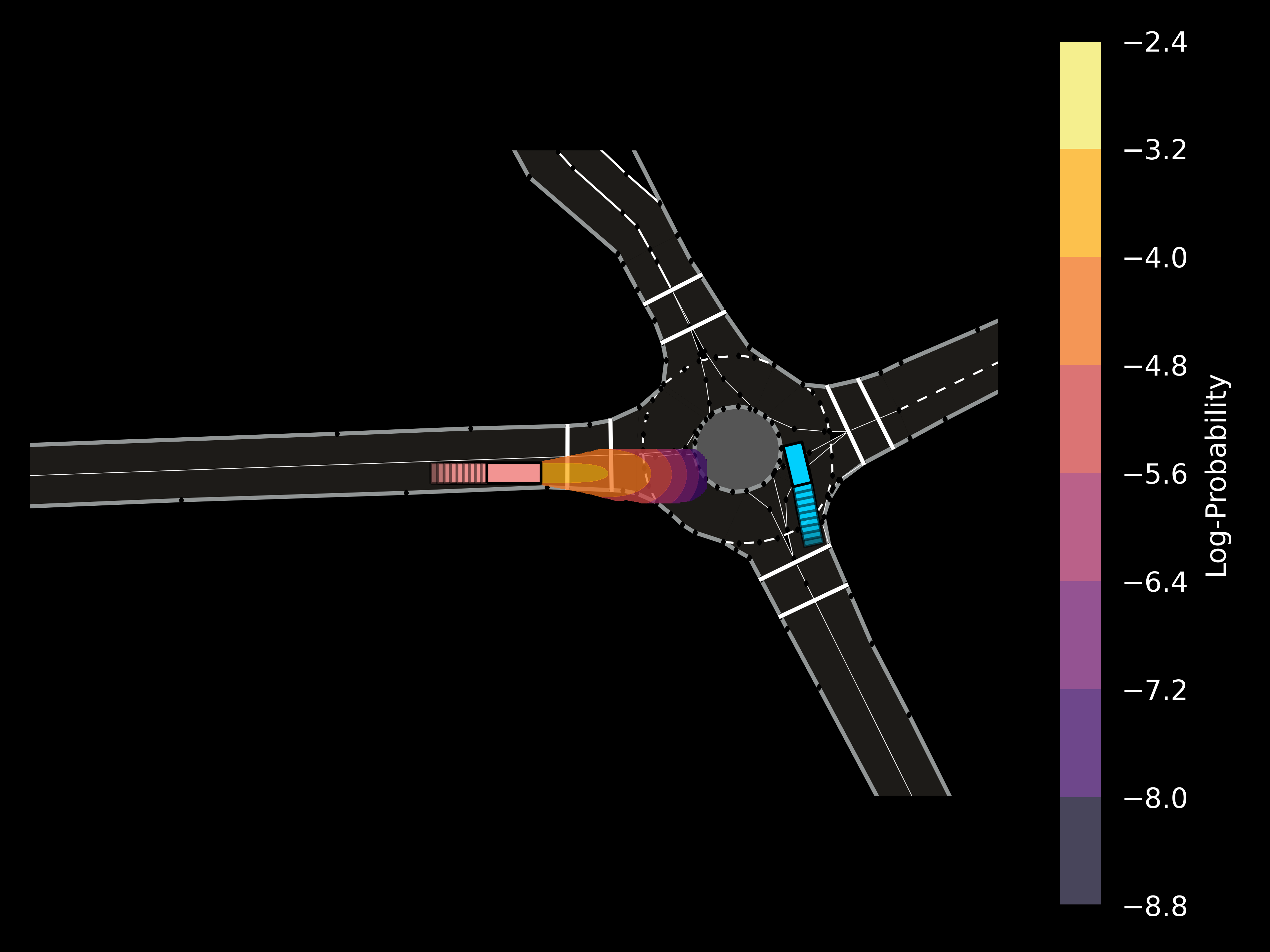}
    \caption{Out-of-Distribution Case (DR\_Roundabout\_RW)}
\end{subfigure}
\caption{\textbf{Example visualization of quantified uncertainty.} We visualize the predicted uncertainty in an in-distribution test case (\textbf{left}) and an OOD test case (\textbf{right}). The heatmap reflects the log-likelihood of a location on the map being visited in the future.}
\label{fig: qualitative}

\end{figure*}

In Figure~\ref{fig: qualitative}, we visualize the quantified uncertainty in an in-distribution and an OOD case side-by-side. The results show that the SeNeVA model can predict a distribution that conforms well to the road geometry even in both in-distribution and OOD cases. The likelihood of a location being visited in the future gradually decreases with respect to the distance, indicating an increased uncertainty in prediction.

\subsection{Motion Prediction}
\label{sec: motion-prediction}

We comprehensively evaluate our SeNeVA model on the INTERACTION and Argoverse 2 dataset motion prediction tasks, comparing its performance with several state-of-the-art motion prediction algorithms. Our results are presented in Table \ref{tab:sota}, demonstrating that the SeNeVA model consistently outperforms other models in the literature. On the INTERACTION dataset, we report our results on the online INTERACTION test set. 
Our SeNeVA model has demonstrated a remarkable enhancement in $\text{minFDE}_6$, outperforming other baseline models by at least 6.8\%. Regarding $\text{minADE}_6$, our SeNeVA model closely approaches the state-of-the-art performance, achieving nearly identical results. Regarding $\text{MR}_6$, SeNeVA ranks as the second-best performing algorithm, with only a marginal 0.21\% difference from the top-performing result. On the INTERACTION val set, all of the metrics outperform the other baselines, with our model achieving a remarkable 35.7\% improvement over the second-best results in $\text{minFDE}_6$ and a substantial 7.0\% lead in $\text{minADE}_6$. Similarly, on the Argoverse 2 validation set, SeNeVA consistently demonstrates competitive results. It achieves a significant 6.8\% lead in $\text{minFDE}_6$ and a remarkable 26.0\% improvement in $\text{minADE}_6$. Additionally, our model's consistently superior performance across various datasets highlights its reliability in handling diverse different scenarios, enhancing safety and efficiency. Our model's lightweight design with just $1.3$ million parameters, shows its efficiency in computational resources and deployment.

\subsection{Ablation Study}
\label{sec: ablation}
We conduct ablation studies to assess the influence of different modules. These experiments are carried out using the validation split of the INTERACTION dataset.

\paragraph{Number of Mixtures} We compare prediction performance varying number of mixtures $N_\text{components}$ in our model. As shown in Table~\ref{tab:ab_mix}, we discover that the optimal choice for the number of mixtures is $6$. We argue that a lower $N_\text{components}$ can limit the expressiveness for multi-modality, while a higher value can lead to a risk of excessive complexity that prevents effective learning. In particular, it affects training the assignment network since the similarities among mixture components increase with increasing mixtures, making it hard for the assignment network to distinguish different distributions.

\paragraph{Assignment Network and NMS Sampling Method} In Table~\ref{tab:ab_an_samp}, we analyze the effect of assignment network (AN) and the sampling method (NMS). The results show that the sampling guided by the assignment network can improve the overall performance. Applying NMS sampling can reduce the MR at the cost of increasing minADE.

\begin{table}[t!]
\centering
\caption{Ablation study investigating the influence of the number of mixture components $N_\text{components}$ in the SeNeVA model.}
\begin{tabular*}{\linewidth}{@{\extracolsep{\fill}}c|ccc@{\extracolsep{\fill}}}
\toprule
$N_\text{components}$ & minFDE$_6 $  & minADE$_6 $ & MR \\
\midrule
4 & 0.5362 & 0.4418 & 0.1065 \\
6 & 0.4306 & 0.1967 & 0.0790 \\
16 & 0.5352 & 0.2030 & 0.1280 \\
32 & 0.5504 & 0.2115 & 0.1316 \\
\bottomrule
\end{tabular*}
\label{tab:ab_mix}
\end{table}

\begin{table}[t!]
\centering
\caption{The performance of SeNeVA on the validation set with and without the inclusion of the assignment network (AN) and NMS sampling method (NMS).}
\begin{tabular*}{\linewidth}{@{\extracolsep{\fill}}ccc|ccc@{\extracolsep{\fill}}}
    \toprule
    \multicolumn{3}{c}{Module} \vrule & \multirow{2}{*}{minFDE$_6$}   & \multirow{2}{*}{minADE$_6$} & \multirow{2}{*}{MR}\\
    \cmidrule{1-3}
    Means & AN & NMS&\\
    \midrule
    \cmark &  &   & 0.5135 & 0.2281 & 0.083 \\
    \cmark & \cmark &  \  & 0.4306 & 0.1967 & 0.079 \\
     & \cmark & \cmark & 0.4265 & 0.2186 & 0.073 \\
    \bottomrule
\end{tabular*}
\label{tab:ab_an_samp}
\end{table}

\section{Conclusion}
\label{sec: conclusion}

This paper introduces SeNeVA, a novel variational Bayes model for uncertainty quantification in motion prediction. An assignment network and an NMS-based trajectory sampling are introduced to support use cases requiring only representative trajectories. Experiments demonstrate SeNeVA's ability to distinguish in-distribution and OOD data by quantifying uncertainty while performing comparative motion prediction to state-of-the-art methods.

\paragraph{Limitations.} SeNeVA requires target-centric inputs and only predicts the distribution of future trajectories for a single agent at a time. Predictions in large-scale traffic scenarios may require parallel inference of multiple models, which can be computationally expensive. Therefore, predicting the joint distribution of multiple agents can be an important future topic to explore.

\vfill

{
    \small
    \bibliographystyle{ieeenat_fullname}
    \bibliography{bibtex/can, bibtex/juanwu}

\begin{thebibliography}{47}
\providecommand{\natexlab}[1]{#1}
\providecommand{\url}[1]{\texttt{#1}}
\expandafter\ifx\csname urlstyle\endcsname\relax
  \providecommand{\doi}[1]{doi: #1}\else
  \providecommand{\doi}{doi: \begingroup \urlstyle{rm}\Url}\fi

\bibitem[Abdar et~al.(2021)Abdar, Pourpanah, Hussain, Rezazadegan, Liu, Ghavamzadeh, Fieguth, Cao, Khosravi, Acharya, Makarenkov, and Nahavandi]{ABDAR2021243}
Moloud Abdar, Farhad Pourpanah, Sadiq Hussain, Dana Rezazadegan, Li Liu, Mohammad Ghavamzadeh, Paul Fieguth, Xiaochun Cao, Abbas Khosravi, U.~Rajendra Acharya, Vladimir Makarenkov, and Saeid Nahavandi.
\newblock A review of uncertainty quantification in deep learning: Techniques, applications and challenges.
\newblock \emph{Information Fusion}, 76:\penalty0 243--297, 2021.

\bibitem[Alahi et~al.(2016)Alahi, Goel, Ramanathan, Robicquet, Fei-Fei, and Savarese]{alahi2016social}
Alexandre Alahi, Kratarth Goel, Vignesh Ramanathan, Alexandre Robicquet, Li Fei-Fei, and Silvio Savarese.
\newblock Social lstm: Human trajectory prediction in crowded spaces.
\newblock In \emph{Proceedings of the IEEE conference on computer vision and pattern recognition}, pages 961--971, 2016.

\bibitem[Chai et~al.(2020)Chai, Sapp, Bansal, and Anguelov]{pmlr-v100-chai20a}
Yuning Chai, Benjamin Sapp, Mayank Bansal, and Dragomir Anguelov.
\newblock Multipath: Multiple probabilistic anchor trajectory hypotheses for behavior prediction.
\newblock In \emph{Proceedings of the Conference on Robot Learning}, pages 86--99. PMLR, 2020.

\bibitem[Chang et~al.(2019)Chang, Lambert, Sangkloy, Singh, Bak, Hartnett, Wang, Carr, Lucey, Ramanan, et~al.]{chang2019argoverse}
Ming-Fang Chang, John Lambert, Patsorn Sangkloy, Jagjeet Singh, Slawomir Bak, Andrew Hartnett, De Wang, Peter Carr, Simon Lucey, Deva Ramanan, et~al.
\newblock Argoverse: 3d tracking and forecasting with rich maps.
\newblock In \emph{Proceedings of the IEEE/CVF conference on computer vision and pattern recognition}, pages 8748--8757, 2019.

\bibitem[Cheng et~al.(2023)Cheng, Mei, and Liu]{cheng_forecast-mae_nodate}
Jie Cheng, Xiaodong Mei, and Ming Liu.
\newblock Forecast-mae: Self-supervised pre-training for motion forecasting with masked autoencoders.
\newblock In \emph{Proceedings of the IEEE/CVF International Conference on Computer Vision (ICCV)}, pages 8679--8689, 2023.

\bibitem[Chiu et~al.(2021)Chiu, Li, Ambru{\c{s}}, and Bohg]{chiu2021probabilistic}
Hsu-kuang Chiu, Jie Li, Rare{\c{s}} Ambru{\c{s}}, and Jeannette Bohg.
\newblock Probabilistic 3d multi-modal, multi-object tracking for autonomous driving.
\newblock In \emph{2021 IEEE international conference on robotics and automation (ICRA)}, pages 14227--14233. IEEE, 2021.

\bibitem[Djuric et~al.(2020)Djuric, Radosavljevic, Cui, Nguyen, Chou, Lin, SINGH, and Schneider]{Djuric_2020_WACV}
Nemanja Djuric, Vladan Radosavljevic, Henggang Cui, Thi Nguyen, Fang-Chieh Chou, Tsung-Han Lin, NITIN SINGH, and Jeff Schneider.
\newblock Uncertainty-aware short-term motion prediction of traffic actors for autonomous driving.
\newblock In \emph{Proceedings of the IEEE/CVF Winter Conference on Applications of Computer Vision (WACV)}, 2020.

\bibitem[Feng et~al.(2021)Feng, Harakeh, Waslander, and Dietmayer]{feng2021review}
Di Feng, Ali Harakeh, Steven~L Waslander, and Klaus Dietmayer.
\newblock A review and comparative study on probabilistic object detection in autonomous driving.
\newblock \emph{IEEE Transactions on Intelligent Transportation Systems}, 23\penalty0 (8):\penalty0 9961--9980, 2021.

\bibitem[Gao et~al.(2020)Gao, Sun, Zhao, Shen, Anguelov, Li, and Schmid]{Gao_2020_CVPR}
Jiyang Gao, Chen Sun, Hang Zhao, Yi Shen, Dragomir Anguelov, Congcong Li, and Cordelia Schmid.
\newblock Vectornet: Encoding hd maps and agent dynamics from vectorized representation.
\newblock In \emph{Proceedings of the IEEE/CVF Conference on Computer Vision and Pattern Recognition (CVPR)}, 2020.

\bibitem[Gilles et~al.(2022)Gilles, Sabatini, Tsishkou, Stanciulescu, and Moutarde]{gilles_gohome_2022}
Thomas Gilles, Stefano Sabatini, Dzmitry Tsishkou, Bogdan Stanciulescu, and Fabien Moutarde.
\newblock {GOHOME}: {Graph}-{Oriented} {Heatmap} {Output} for future {Motion} {Estimation}.
\newblock In \emph{2022 {International} {Conference} on {Robotics} and {Automation} ({ICRA})}, pages 9107--9114, Philadelphia, PA, USA, 2022. IEEE.

\bibitem[Goodfellow et~al.(2014)Goodfellow, Pouget-Abadie, Mirza, Xu, Warde-Farley, Ozair, Courville, and Bengio]{goodfellow2014generative}
Ian Goodfellow, Jean Pouget-Abadie, Mehdi Mirza, Bing Xu, David Warde-Farley, Sherjil Ozair, Aaron Courville, and Yoshua Bengio.
\newblock Generative adversarial nets.
\newblock \emph{Advances in neural information processing systems}, 27, 2014.

\bibitem[Gu et~al.(2021)Gu, Sun, and Zhao]{densetnt}
Junru Gu, Chen Sun, and Hang Zhao.
\newblock Densetnt: End-to-end trajectory prediction from dense goal sets.
\newblock In \emph{Proceedings of the IEEE/CVF International Conference on Computer Vision}, pages 15303--15312, 2021.

\bibitem[Gupta et~al.(2018)Gupta, Johnson, Fei-Fei, Savarese, and Alahi]{gupta2018social}
Agrim Gupta, Justin Johnson, Li Fei-Fei, Silvio Savarese, and Alexandre Alahi.
\newblock Social gan: Socially acceptable trajectories with generative adversarial networks.
\newblock In \emph{Proceedings of the IEEE conference on computer vision and pattern recognition}, pages 2255--2264, 2018.

\bibitem[Hall et~al.(2020)Hall, Dayoub, Skinner, Zhang, Miller, Corke, Carneiro, Angelova, and S{\"u}nderhauf]{hall2020probabilistic}
David Hall, Feras Dayoub, John Skinner, Haoyang Zhang, Dimity Miller, Peter Corke, Gustavo Carneiro, Anelia Angelova, and Niko S{\"u}nderhauf.
\newblock Probabilistic object detection: Definition and evaluation.
\newblock In \emph{Proceedings of the IEEE/CVF Winter Conference on Applications of Computer Vision}, pages 1031--1040, 2020.

\bibitem[Hosang et~al.(2017)Hosang, Benenson, and Schiele]{hosang2017learning}
Jan Hosang, Rodrigo Benenson, and Bernt Schiele.
\newblock Learning non-maximum suppression.
\newblock In \emph{Proceedings of the IEEE conference on computer vision and pattern recognition}, pages 4507--4515, 2017.

\bibitem[Huang et~al.(2022)Huang, Mo, and Lv]{huang_multi-modal_2022}
Zhiyu Huang, Xiaoyu Mo, and Chen Lv.
\newblock Multi-modal {Motion} {Prediction} with {Transformer}-based {Neural} {Network} for {Autonomous} {Driving}.
\newblock In \emph{2022 {International} {Conference} on {Robotics} and {Automation} ({ICRA})}, pages 2605--2611, Philadelphia, PA, USA, 2022. IEEE.

\bibitem[Ivanovic and Pavone(2019)]{ivanovic2019trajectron}
Boris Ivanovic and Marco Pavone.
\newblock The trajectron: Probabilistic multi-agent trajectory modeling with dynamic spatiotemporal graphs.
\newblock In \emph{Proceedings of the IEEE/CVF International Conference on Computer Vision}, pages 2375--2384, 2019.

\bibitem[Ivanovic et~al.(2018)Ivanovic, Schmerling, Leung, and Pavone]{ivanovic2018generative}
Boris Ivanovic, Edward Schmerling, Karen Leung, and Marco Pavone.
\newblock Generative modeling of multimodal multi-human behavior.
\newblock In \emph{2018 IEEE/RSJ International Conference on Intelligent Robots and Systems (IROS)}, pages 3088--3095. IEEE, 2018.

\bibitem[Janjo{\v s} et~al.(2023)Janjo{\v s}, Keller, Dolgov, and Z{\"o}llner]{janjos_bridging_2023}
Faris Janjo{\v s}, Max Keller, Maxim Dolgov, and J.~Marius Z{\"o}llner.
\newblock Bridging the {Gap} {Between} {Multi}-{Step} and {One}-{Shot} {Trajectory} {Prediction} via {Self}-{Supervision}.
\newblock In \emph{2023 {IEEE} {Intelligent} {Vehicles} {Symposium} ({IV})}, pages 1--8, Anchorage, AK, USA, 2023. IEEE.

\bibitem[Jia et~al.(2022)Jia, Wu, Chen, Li, Liu, and Yan]{jia_hdgt_2022}
Xiaosong Jia, Penghao Wu, Li Chen, Hongyang Li, Yu Liu, and Junchi Yan.
\newblock {HDGT}: {Heterogeneous} {Driving} {Graph} {Transformer} for {Multi}-{Agent} {Trajectory} {Prediction} via {Scene} {Encoding}, 2022.
\newblock arXiv:2205.09753 [cs].

\bibitem[Kingma and Welling(2022)]{kingma2022autoencoding}
Diederik~P Kingma and Max Welling.
\newblock Auto-encoding variational bayes, 2022.

\bibitem[Kodali et~al.(2017)Kodali, Abernethy, Hays, and Kira]{kodali2017convergence}
Naveen Kodali, Jacob Abernethy, James Hays, and Zsolt Kira.
\newblock On convergence and stability of gans, 2017.

\bibitem[Kosaraju et~al.(2019)Kosaraju, Sadeghian, Mart{\'\i}n-Mart{\'\i}n, Reid, Rezatofighi, and Savarese]{kosaraju2019social}
Vineet Kosaraju, Amir Sadeghian, Roberto Mart{\'\i}n-Mart{\'\i}n, Ian Reid, Hamid Rezatofighi, and Silvio Savarese.
\newblock Social-bigat: Multimodal trajectory forecasting using bicycle-gan and graph attention networks.
\newblock \emph{Advances in Neural Information Processing Systems}, 32, 2019.

\bibitem[Kullback and Leibler(1951)]{10.1214/aoms/1177729694}
S. Kullback and R.~A. Leibler.
\newblock {On Information and Sufficiency}.
\newblock \emph{The Annals of Mathematical Statistics}, 22\penalty0 (1):\penalty0 79 -- 86, 1951.

\bibitem[Lakshminarayanan et~al.(2017)Lakshminarayanan, Pritzel, and Blundell]{NIPS2017_9ef2ed4b}
Balaji Lakshminarayanan, Alexander Pritzel, and Charles Blundell.
\newblock Simple and scalable predictive uncertainty estimation using deep ensembles.
\newblock In \emph{Advances in Neural Information Processing Systems}. Curran Associates, Inc., 2017.

\bibitem[Lee et~al.(2017)Lee, Choi, Vernaza, Choy, Torr, and Chandraker]{lee_desire_2017}
Namhoon Lee, Wongun Choi, Paul Vernaza, Christopher~B. Choy, Philip H.~S. Torr, and Manmohan Chandraker.
\newblock {DESIRE}: {Distant} {Future} {Prediction} in {Dynamic} {Scenes} with {Interacting} {Agents}.
\newblock In \emph{2017 {IEEE} {Conference} on {Computer} {Vision} and {Pattern} {Recognition} ({CVPR})}, pages 2165--2174, Honolulu, HI, 2017. IEEE.

\bibitem[Liang et~al.(2020)Liang, Yang, Hu, Chen, Liao, Feng, and Urtasun]{liang2020learning}
Ming Liang, Bin Yang, Rui Hu, Yun Chen, Renjie Liao, Song Feng, and Raquel Urtasun.
\newblock Learning lane graph representations for motion forecasting.
\newblock In \emph{Computer Vision--ECCV 2020: 16th European Conference, Glasgow, UK, August 23--28, 2020, Proceedings, Part II 16}, pages 541--556. Springer, 2020.

\bibitem[Liao et~al.(2022)Liao, Wang, Zhao, Zhao, Han, Tiwari, Barth, and Wu]{liao2022online}
Xishun Liao, Ziran Wang, Xuanpeng Zhao, Zhouqiao Zhao, Kyungtae Han, Prashant Tiwari, Matthew~J Barth, and Guoyuan Wu.
\newblock Online prediction of lane change with a hierarchical learning-based approach.
\newblock In \emph{2022 IEEE International Conference on Robotics and Automation (ICRA)}, 2022.

\bibitem[Lin et~al.(2017)Lin, Goyal, Girshick, He, and Dollar]{Lin_2017_ICCV}
Tsung-Yi Lin, Priya Goyal, Ross Girshick, Kaiming He, and Piotr Dollar.
\newblock Focal loss for dense object detection.
\newblock In \emph{Proceedings of the IEEE International Conference on Computer Vision (ICCV)}, 2017.

\bibitem[Liu et~al.(2021)Liu, Zhang, Fang, Jiang, and Zhou]{liu_multimodal_2021}
Yicheng Liu, Jinghuai Zhang, Liangji Fang, Qinhong Jiang, and Bolei Zhou.
\newblock Multimodal {Motion} {Prediction} with {Stacked} {Transformers}.
\newblock In \emph{2021 {IEEE}/{CVF} {Conference} on {Computer} {Vision} and {Pattern} {Recognition} ({CVPR})}, pages 7573--7582, Nashville, TN, USA, 2021. IEEE.

\bibitem[Liu et~al.(2022)Liu, Wang, Han, Shou, Tiwari, and Hansen]{liu2022vision}
Yongkang Liu, Ziran Wang, Kyungtae Han, Zhenyu Shou, Prashant Tiwari, and John Hansen.
\newblock Vision-cloud data fusion for adas: A lane change prediction case study.
\newblock \emph{IEEE Transactions on Intelligent Vehicles}, 7\penalty0 (2):\penalty0 210--220, 2022.

\bibitem[Lu et~al.(2024)Lu, Zhan, Tomizuka, and Hu]{lu2024generalizable}
Juanwu Lu, Wei Zhan, Masayoshi Tomizuka, and Yeping Hu.
\newblock Towards generalizable and interpretable motion prediction: A deep variational bayes approach, 2024.

\bibitem[Memory(2010)]{memory2010long}
Long Short-Term Memory.
\newblock Long short-term memory.
\newblock \emph{Neural computation}, 9\penalty0 (8):\penalty0 1735--1780, 2010.

\bibitem[Morton et~al.(2016)Morton, Wheeler, and Kochenderfer]{morton2016analysis}
Jeremy Morton, Tim~A Wheeler, and Mykel~J Kochenderfer.
\newblock Analysis of recurrent neural networks for probabilistic modeling of driver behavior.
\newblock \emph{IEEE Transactions on Intelligent Transportation Systems}, 18\penalty0 (5):\penalty0 1289--1298, 2016.

\bibitem[Neubeck and Van~Gool(2006)]{neubeck2006efficient}
Alexander Neubeck and Luc Van~Gool.
\newblock Efficient non-maximum suppression.
\newblock In \emph{18th international conference on pattern recognition (ICPR'06)}, pages 850--855. IEEE, 2006.

\bibitem[Sadeghian et~al.(2019)Sadeghian, Kosaraju, Sadeghian, Hirose, Rezatofighi, and Savarese]{sadeghian2019sophie}
Amir Sadeghian, Vineet Kosaraju, Ali Sadeghian, Noriaki Hirose, Hamid Rezatofighi, and Silvio Savarese.
\newblock Sophie: An attentive gan for predicting paths compliant to social and physical constraints.
\newblock In \emph{Proceedings of the IEEE/CVF conference on computer vision and pattern recognition}, pages 1349--1358, 2019.

\bibitem[Salzmann et~al.(2020)Salzmann, Ivanovic, Chakravarty, and Pavone]{salzmann2020trajectron++}
Tim Salzmann, Boris Ivanovic, Punarjay Chakravarty, and Marco Pavone.
\newblock Trajectron++: Dynamically-feasible trajectory forecasting with heterogeneous data.
\newblock In \emph{Computer Vision--ECCV 2020: 16th European Conference, Glasgow, UK, August 23--28, 2020, Proceedings, Part XVIII 16}, pages 683--700. Springer, 2020.

\bibitem[Schmerling et~al.(2018)Schmerling, Leung, Vollprecht, and Pavone]{schmerling2018multimodal}
Edward Schmerling, Karen Leung, Wolf Vollprecht, and Marco Pavone.
\newblock Multimodal probabilistic model-based planning for human-robot interaction.
\newblock In \emph{2018 IEEE International Conference on Robotics and Automation (ICRA)}, pages 3399--3406. IEEE, 2018.

\bibitem[Sohn et~al.(2015)Sohn, Lee, and Yan]{sohn2015learning}
Kihyuk Sohn, Honglak Lee, and Xinchen Yan.
\newblock Learning structured output representation using deep conditional generative models.
\newblock \emph{Advances in neural information processing systems}, 28, 2015.

\bibitem[Sun et~al.(2020)Sun, Kretzschmar, Dotiwalla, Chouard, Patnaik, Tsui, Guo, Zhou, Chai, Caine, et~al.]{sun2020scalability}
Pei Sun, Henrik Kretzschmar, Xerxes Dotiwalla, Aurelien Chouard, Vijaysai Patnaik, Paul Tsui, James Guo, Yin Zhou, Yuning Chai, Benjamin Caine, et~al.
\newblock Scalability in perception for autonomous driving: Waymo open dataset.
\newblock In \emph{Proceedings of the IEEE/CVF conference on computer vision and pattern recognition}, pages 2446--2454, 2020.

\bibitem[Vemula et~al.(2018)Vemula, Muelling, and Oh]{vemula2018social}
Anirudh Vemula, Katharina Muelling, and Jean Oh.
\newblock Social attention: Modeling attention in human crowds.
\newblock In \emph{2018 IEEE international Conference on Robotics and Automation (ICRA)}, pages 4601--4607. IEEE, 2018.

\bibitem[Wang et~al.(2008)Wang, Fleet, and Hertzmann]{4359316}
Jack~M. Wang, David~J. Fleet, and Aaron Hertzmann.
\newblock Gaussian process dynamical models for human motion.
\newblock \emph{IEEE Transactions on Pattern Analysis and Machine Intelligence}, 30\penalty0 (2):\penalty0 283--298, 2008.

\bibitem[Wang et~al.(2021)Wang, Pang, Chen, Iyer, Dutta, Menon, and Liu]{WANG2021107650}
Yuhao Wang, Yutian Pang, Oliver Chen, Hari~N. Iyer, Parikshit Dutta, P.K. Menon, and Yongming Liu.
\newblock Uncertainty quantification and reduction in aircraft trajectory prediction using bayesian-entropy information fusion.
\newblock \emph{Reliability Engineering \& System Safety}, 212:\penalty0 107650, 2021.

\bibitem[Wilson et~al.(2023)Wilson, Qi, Agarwal, Lambert, Singh, Khandelwal, Pan, Kumar, Hartnett, Pontes, Ramanan, Carr, and Hays]{wilson2023argoverse}
Benjamin Wilson, William Qi, Tanmay Agarwal, John Lambert, Jagjeet Singh, Siddhesh Khandelwal, Bowen Pan, Ratnesh Kumar, Andrew Hartnett, Jhony~Kaesemodel Pontes, Deva Ramanan, Peter Carr, and James Hays.
\newblock Argoverse 2: Next generation datasets for self-driving perception and forecasting, 2023.

\bibitem[Zhan et~al.(2019)Zhan, Sun, Wang, Shi, Clausse, Naumann, Kummerle, Konigshof, Stiller, de~La~Fortelle, and Tomizuka]{zhan2019interaction}
Wei Zhan, Liting Sun, Di Wang, Haojie Shi, Aubrey Clausse, Maximilian Naumann, Julius Kummerle, Hendrik Konigshof, Christoph Stiller, Arnaud de La~Fortelle, and Masayoshi Tomizuka.
\newblock Interaction dataset: An international, adversarial and cooperative motion dataset in interactive driving scenarios with semantic maps, 2019.

\bibitem[Zhao et~al.(2021)Zhao, Gao, Lan, Sun, Sapp, Varadarajan, Shen, Shen, Chai, Schmid, Li, and Anguelov]{zhao_tnt_nodate}
Hang Zhao, Jiyang Gao, Tian Lan, Chen Sun, Ben Sapp, Balakrishnan Varadarajan, Yue Shen, Yi Shen, Yuning Chai, Cordelia Schmid, Congcong Li, and Dragomir Anguelov.
\newblock Tnt: Target-driven trajectory prediction.
\newblock In \emph{Proceedings of the 2020 Conference on Robot Learning}, pages 895--904. PMLR, 2021.

\bibitem[Zhao et~al.(2019)Zhao, Xu, Monfort, Choi, Baker, Zhao, Wang, and Wu]{zhao2019multi}
Tianyang Zhao, Yifei Xu, Mathew Monfort, Wongun Choi, Chris Baker, Yibiao Zhao, Yizhou Wang, and Ying~Nian Wu.
\newblock Multi-agent tensor fusion for contextual trajectory prediction.
\newblock In \emph{Proceedings of the IEEE/CVF Conference on Computer Vision and Pattern Recognition}, pages 12126--12134, 2019.

\end{thebibliography}
}

\clearpage
\appendix
\setcounter{page}{1}
\maketitlesupplementary

\section{Implementation Details}
\label{sec: supp-implementation-detail}

 We implement our model using PyTorch, trained for $20$ epochs on the INTERACTION dataset with a batch size of $64$ and $25$ epochs on the Argoverse 2 dataset with a batch size of $64$. With only $1.3$M parameters, the model balances scalability and performance. We set $\alpha=1$ and use the Adam optimization solver with a learning rate of $0.0001$ and the learning rate decay schedule with a step size of $5$ epochs and a rate of $0.3$ to ensure efficient convergence. We train and evaluate our model using only a single NVIDIA GeForce RTX 3090 Ti.

\section{Evaluation Metrics}
\label{sec: supp-metrics}

\subsection{Uncertainty Quantification}

In our formulation, the random variables $\boldsymbol{s}_\text{f}$ and $\boldsymbol{v}$ are both Gaussian variables, and $z$ follows a categorical distribution. Therefore, we can compute the total uncertainty for a predicted distribution by its entropy. Given the generative model in equation~\ref{eq: generative-model}, the total entropy can be estimated by the summation of three individual expected entropy:
\begin{equation}
\begin{aligned}
    & \sum_z\int_{\boldsymbol{v}}\int_{\boldsymbol{s}_\text{f}}p(\boldsymbol{s}_\text{f},\boldsymbol{v},z|\boldsymbol{x})\log p(\boldsymbol{s}_\text{f},\boldsymbol{v},z|\boldsymbol{x})d{\boldsymbol{s}_\text{f}}d\boldsymbol{v} \\
    = & \mathbb{E}_{\boldsymbol{v},z\sim p(\boldsymbol{v},z|\boldsymbol{x})}\text{Entropy}(p(\boldsymbol{s}_\text{f}|\boldsymbol{v},x)) \\
    & + \mathbb{E}_{\boldsymbol{s}_\text{f}\sim p(z\sim p(z)}\text{Entropy}\left(p(\boldsymbol{v}|\boldsymbol{x},z)\right) \\
    & + \text{Entropy}\left(p(z)\right).
\end{aligned}
\end{equation}
Since we have a fixed prior $p(z)$, the comparison of the total entropy reduces to comparing the sum of the first two terms. In our experiment, we use Monte-Carlo sampling to generate $N_\text{mc}$ samples of $\boldsymbol{v}$ for entropy calculation.

\subsection{Motion Prediction}
For motion prediction, we use the standard Minimum Average Displacement Error (minADE), Minimum Final Displacement Error (minFDE), and Miss Rate (MR) to assess the accuracy and effectiveness of our approach. minADE and minFDE are distance-based metrics commonly used in multi-modal trajectory prediction (i.e., trajectory prediction with multiple possible outcomes) tasks. The minADE calculates the average Euclidean distance between predicted and ground truth trajectories at each time step, taking the minimum across all trajectories in the prediction set:
\begin{equation}
    \text{minADE}(\hat{x}_n^k, x_n) = \frac{1}{NT}\sum\limits_{n=1}^N\min\limits_{k=1,\ldots, K}\sum\limits_{t=1}^T\left\|\hat{x}_{n, t}^k - x_{n, t}\right\|_2.
\end{equation}
On the other hand, the minFDE measures the Euclidean distance between predicted and ground truth final positions, effectively assessing the long-term prediction performance of the model:
\begin{equation}
    \text{minFDE}(\hat{x}_n^k, x_n)=\frac{1}{N}\sum\limits_{n=1}^N\min\limits_{k=1,\ldots,K}\left\|\hat{x}_{n, T}^k-x_{n, T}\right\|_2.
\end{equation}

MR represents the ratio of 'miss' cases over all cases. The definitions of MR are significantly different for the INTERACTION dataset and the Argoverse 2 dataset. 

In the INTERACTION dataset, if its prediction at the final timestamp (T=30) is out of a given lateral or longitudinal threshold of the ground truth, it will be assumed as a 'miss.' In the INTERACTION dataset, we need to align both the ground truth and the prediction by rotating them based on the yaw angle of the ground truth at the final timestamp, ensuring that the x-axis represents the longitudinal direction and the y-axis corresponds to the lateral direction. The lateral threshold is established as 1 meter, while the longitudinal threshold is a piecewise function set as:

\begin{linenomath}
    \begin{equation}
        \text{Threshold}_\text{lon}=
        \begin{cases}
        1 & v<1.4m/s \\
        1 + \frac{v-1.4}{11-1.4} & 1.4m/s \leq v \leq 11m/s\\
        2 & v \geq 11m/s
    \end{cases}
    \end{equation}
\end{linenomath}

For the Argoverse 2 dataset, the MR indicates the proportion of test samples where none of the predicted trajectories fall within a $2$-meter range of the ground truth, as measured through the endpoint error measurement.

\begin{figure*}[t!]
    \centering
    \begin{subfigure}{0.4\textwidth}
        \centering
        \includegraphics[width=\textwidth]{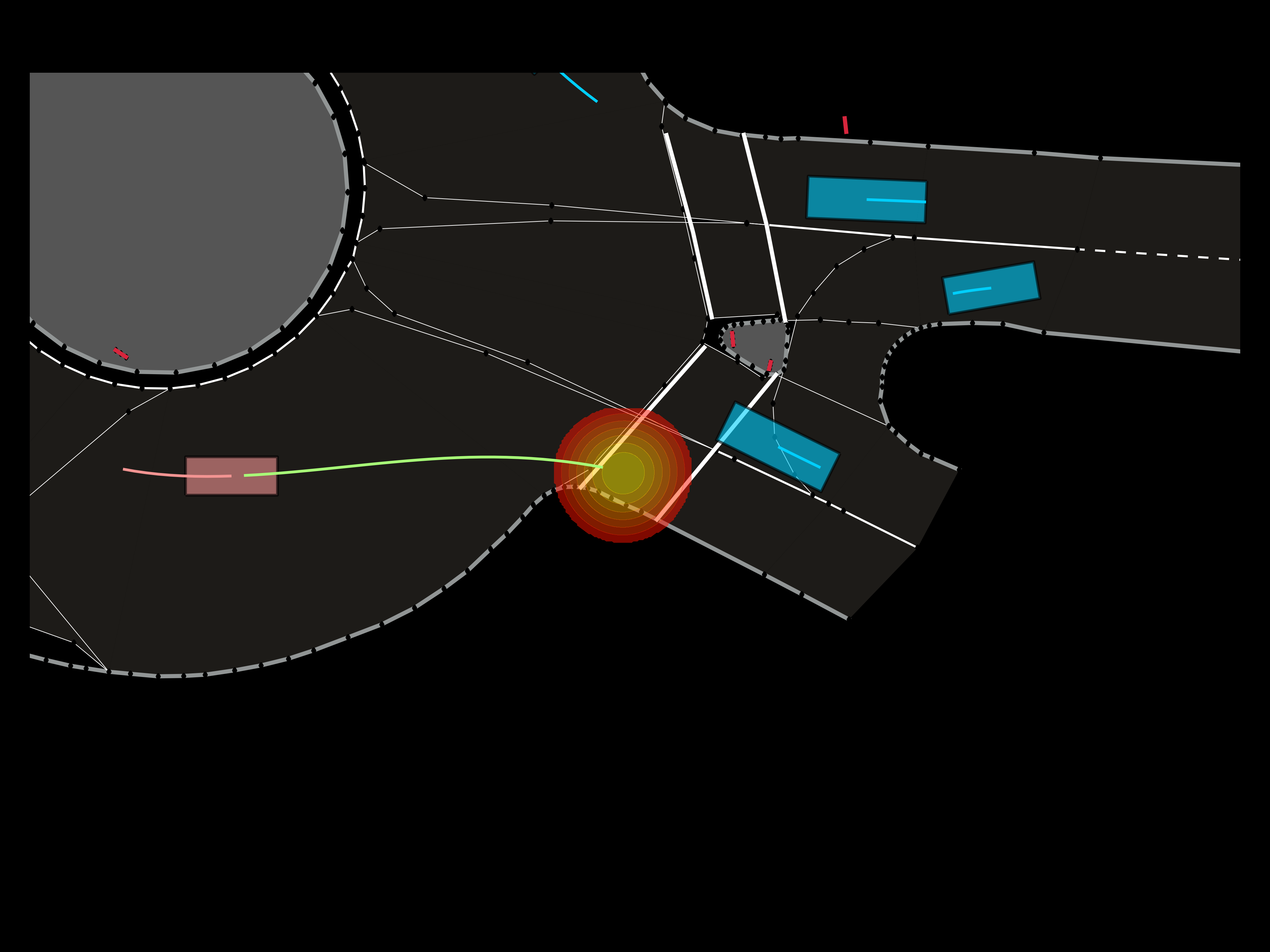}
        \caption{\textbf{Distribution evaluation output.} The heatmap illustrate the distribution of $y_{H+T}$ in this case quantified by the SeNeVA model.}
        \label{fig: evaluation}
    \end{subfigure}
    \hspace{1em}
    \begin{subfigure}{0.4\textwidth}
        \centering
        \includegraphics[width=\textwidth]{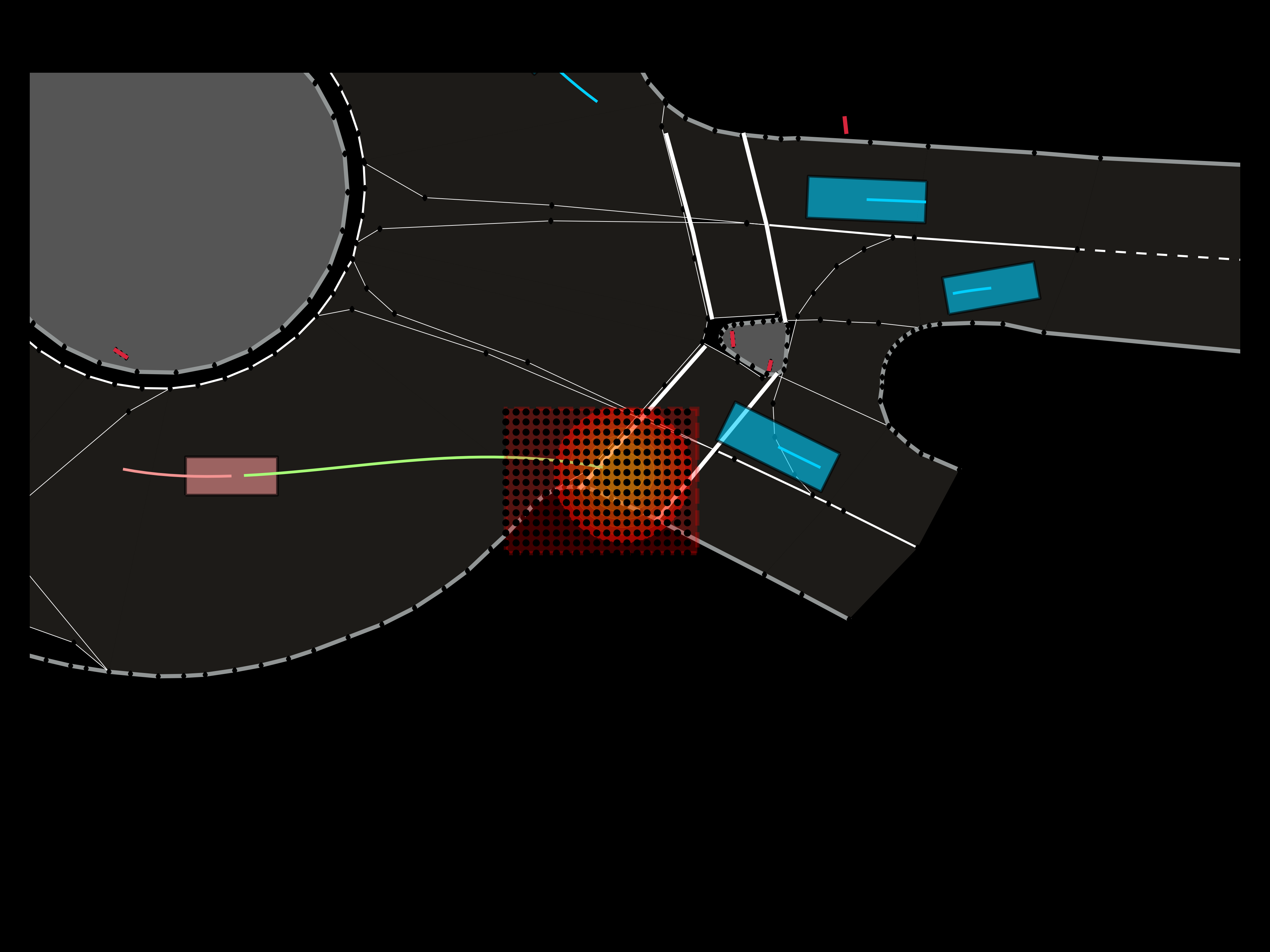}
        \caption{\textbf{Dense grid generation.} For sampling, we generate a grid of candidates that covers $2$ standard deviation area of the $y_{H+T}$ distribution.}
        \label{fig: dense-grid}
    \end{subfigure}
    \vspace{2em}
    \begin{subfigure}{0.4\textwidth}
        \centering
        \includegraphics[width=\textwidth]{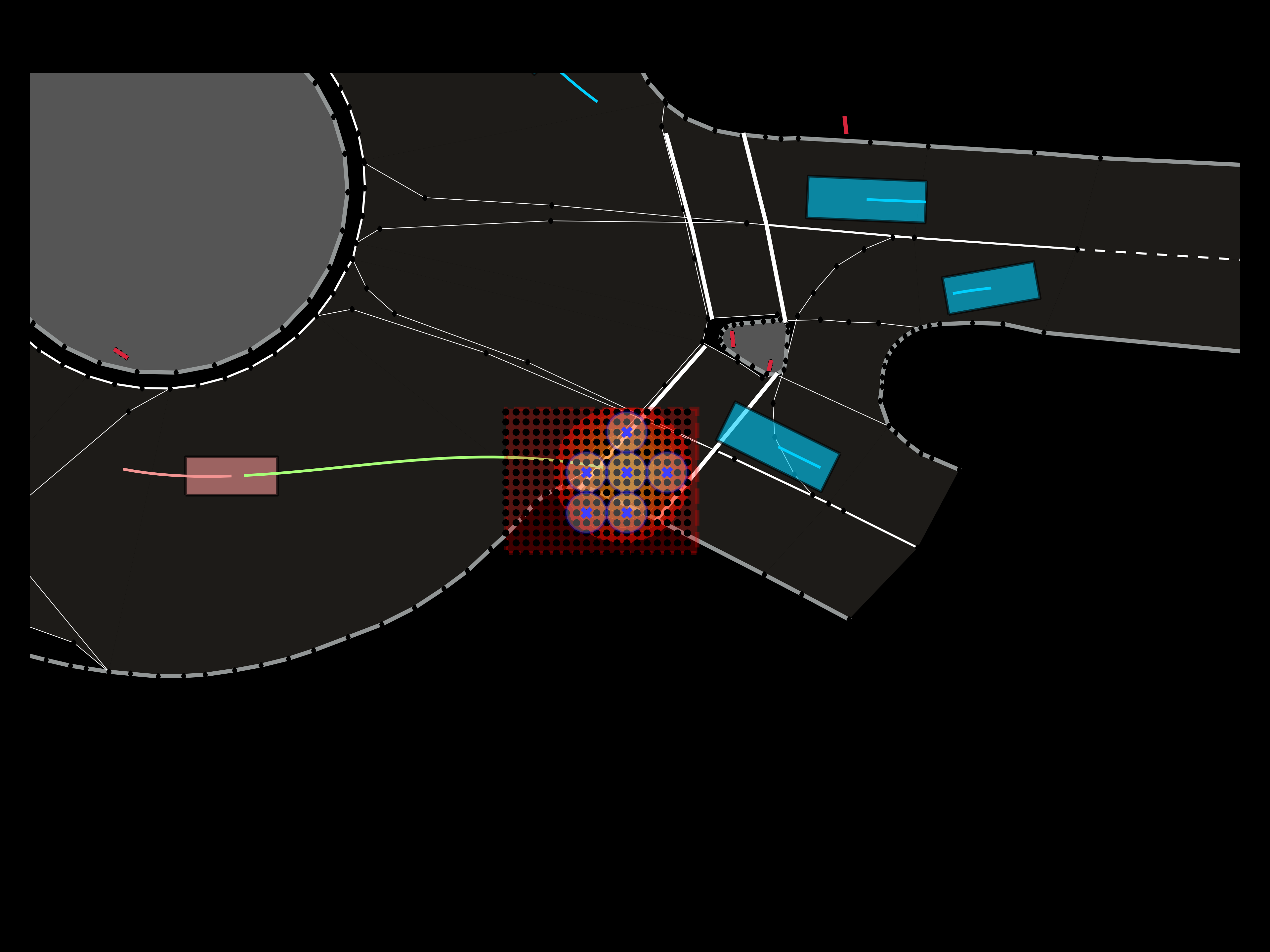}
        \caption{\textbf{NMS sampling results.} Following Algorithm~\ref{alg: sampling}, we can sample a set of top-$M$ candidates (\textcolor{Blue}{blue}) regarding circular buffers defined by radius $r$ and their IoU threshold $\gamma$.}
        \label{fig: nms-sampling}
    \end{subfigure}
    \hspace{1em}
    \begin{subfigure}{0.4\textwidth}
        \centering
        \includegraphics[width=\textwidth]{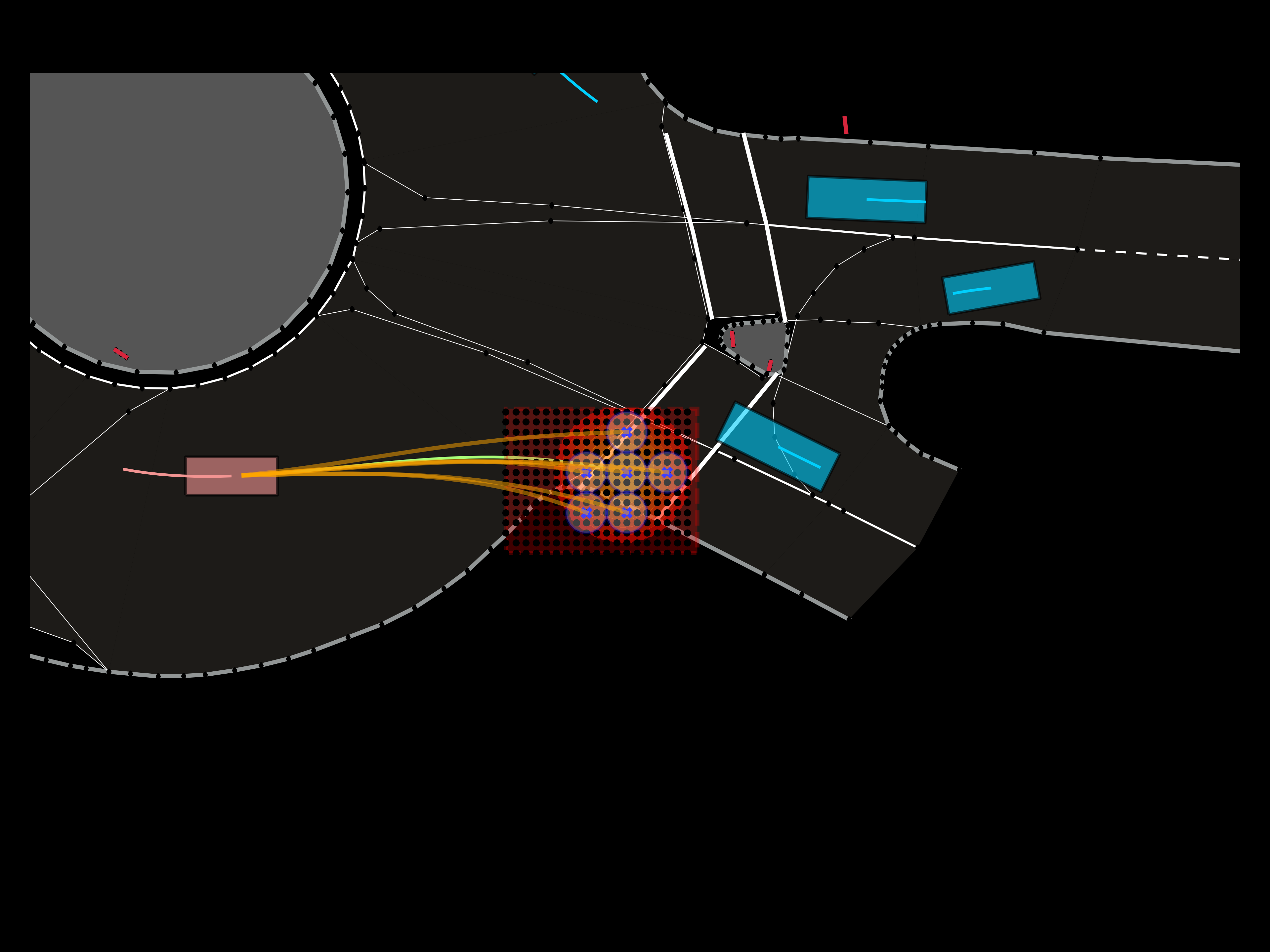}
        \caption{\textbf{Backward completion.} Starting from the last location, we assume homogeneous uncertainty over time and compute intermediate waypoints to obtain the final trajectories (\textcolor{Orange}{orange}).}
        \label{fig: completion}
    \end{subfigure}
    \vspace{-2em}
    \caption{\textbf{Example visualization of the backward sampling process.} The multi-modal trajectory prediction is generated through (a) evaluating distribution, (b) generating dense candidates, (c) applying NMS sampling, and finally (d) completing intermediate trajectories. To better illustrate the effectiveness of our method, we plot the history (\textcolor{Red}{red}) and ground-truth future trajectory (\textcolor{Green}{green}) of the target agent.}
\end{figure*}

\section{Methodology Details}
\label{sec: supp-method}

\subsection{Derivation of Evidence Lower Bound (ELBO)}
\label{sec: elbo-loss}
The standard and intuitive objective for training the probabilistic model in SeNeVA is to let the modeled conditional distribution $p(\boldsymbol{s}_\text{f}|\boldsymbol{x})$ to match ground-truth data distribution through maximizing the likelihood. However, direct computation on the likelihood function is intractable since it involves calculating the integration given as $p(\boldsymbol{s}_\text{f}|\boldsymbol{x})=\sum_z\int_{\boldsymbol{v}}p(\boldsymbol{s}_\text{f},\boldsymbol{v},z|\boldsymbol{x})d\boldsymbol{v}dz$, which is hard to estimate and optimize. To address this issue, we follow the popular \textit{variational inference} method and introduce a tractable, closed-form, and easy-sampling proxy posterior $q(\boldsymbol{v},z|\boldsymbol{s}_\text{f},\boldsymbol{x})$ of the latent variables conditioned on the observed variables, and the lower bound of the log-likelihood can be derived with Jensen's Inequality:
\begin{equation}
\begin{aligned}
    \log p(\boldsymbol{s}_\text{f}|\boldsymbol{x}) &= \log\int_z\int_{\boldsymbol{v}}p(\boldsymbol{s}_\text{f},\boldsymbol{v},z|\boldsymbol{x})d\boldsymbol{v}dz \\
    &= \log\mathbb{E}_{q(\boldsymbol{v},z|\boldsymbol{s}_\text{f},\boldsymbol{x})}\left[\frac{p(\boldsymbol{s}_\text{f},\boldsymbol{v},z|\boldsymbol{x})}{q(\boldsymbol{v},z|\boldsymbol{s}_\text{f},\boldsymbol{x})}\right] \\
    &\geq \mathbb{E}_{q(\boldsymbol{v},z|\boldsymbol{s}_\text{f},\boldsymbol{x})}\log\left[\frac{p(\boldsymbol{s}_\text{f},\boldsymbol{v},z|\boldsymbol{x})}{q(\boldsymbol{v},z|\boldsymbol{s}_\text{f},\boldsymbol{x})}\right].
\end{aligned}
\end{equation}

Since we factorize the joint distribution $p(\boldsymbol{s}_\text{f},\boldsymbol{v},z|\boldsymbol{x})$ and the posterior $q(\boldsymbol{v},z|\boldsymbol{s}_\text{f},\boldsymbol{x})$ in equation~\ref{eq: generative-model} and equation~\ref{eq: variational-family}, respectively, we can leverage the factorization and expand the expectation term to compute the analytical solution of the lower bound. To simplify the following notation, we denote $p(\boldsymbol{s}_\text{f},\boldsymbol{v},z|x) = p_{\boldsymbol{s}_\text{f},\boldsymbol{v},z}$, $q(\boldsymbol{v},z|\boldsymbol{s}_\text{f},\boldsymbol{x}) = q_{\boldsymbol{v},z}$, $p(\boldsymbol{s}_\text{f}|\boldsymbol{v},\boldsymbol{x}) = p_{\boldsymbol{s}_\text{f}}$, $p(\boldsymbol{v}|\boldsymbol{x},z)=p_{\boldsymbol{v}}$, $p(z)=p_z$, $q(\boldsymbol{v}|\boldsymbol{s}_\text{f},\boldsymbol{x})=q_{\boldsymbol{v}}$, and $q(z|\boldsymbol{v},\boldsymbol{x})=q_z$. The expansion of the expectation term writes:
\begin{equation}
\begin{aligned}
    \mathbb{E}_{q_{\boldsymbol{v},z}}&\log\left[\frac{p_{\boldsymbol{s}_\text{f},\boldsymbol{v},z}}{q_{\boldsymbol{v},z}}\right] = \int_z\int_{\boldsymbol{v}} \log\left[\frac{p_{\boldsymbol{s}_\text{f}}p_{\boldsymbol{v}}p_z}{q_{\boldsymbol{v}}q_z}\right]\cdot q_{\boldsymbol{v},z}d\boldsymbol{v}dz \\
    &= \mathbb{E}_{q_{\boldsymbol{v}}}\log p_{\boldsymbol{s}_\text{f}} + \int_zq_z\int_{\boldsymbol{v}}q_{\boldsymbol{v}}\log\left[\frac{p_{\boldsymbol{v}}}{q_{\boldsymbol{v}}}\right]d\boldsymbol{v}dz \\
    &\quad + \int_{\boldsymbol{v}}q_{\boldsymbol{v}}\int_zq_z\log\left[\frac{p_z}{q_z}\right]dzd\boldsymbol{v} \\
    &= \mathbb{E}_{q_{\boldsymbol{v}}}\left(\log p_{\boldsymbol{s}_\text{f}} - D_\text{KL}(q_z\|p_z)\right) - \mathbb{E}_{q_z}D_\text{KL}(q_{\boldsymbol{v}}\|p_{\boldsymbol{v}}).
\end{aligned}
\end{equation}

The expansion above is the ELBO objective we maximize during training equivalent to the formula given in equation~\ref{eq: elbo-expr}. Therefore, maximizing the lower bound is equal to minimizing the KL divergence, driving the variational posterior $q(\boldsymbol{v},z|\boldsymbol{s}_\text{f},\boldsymbol{x})$ towards the ground-truth posterior. As a result, maximizing the ELBO objective can effectively maximize the likelihood.

\subsection{Derivation of Assignment Network Loss}

The assignment network directly approximates $p(z|\boldsymbol{x})$ to avoid tedious sampling at inference time from the latent $\boldsymbol{v}$ space to estimate the posterior $q(z|\boldsymbol{x})=\int_{\boldsymbol{v}}q(z|\boldsymbol{v},\boldsymbol{x})d\boldsymbol{v}$. We can obtain the distribution over $z$ given in equation~\ref{eq: z-proxy-target} by applying Bayes' rule. Herein, we estimate the conditional distribution $p(\boldsymbol{s}_\text{f}|\boldsymbol{x},z)$ by applying Monte-Carlo sampling over the latent $\boldsymbol{v}$ space at training:
\begin{equation}
\begin{aligned}
    p(\boldsymbol{s}_\text{f}|\boldsymbol{x},z) &= \int_{\boldsymbol{v}}p(\boldsymbol{s}_\text{f},\boldsymbol{v}|\boldsymbol{x},z)d\boldsymbol{v} \\
    &\approx\frac{1}{N_\text{mc}}\sum\limits_{n=1}^{N_\text{mc}}p(\boldsymbol{s}_\text{f},\boldsymbol{v}^{(n)}|\boldsymbol{x})p(\boldsymbol{v}^{(n)},z|\boldsymbol{x}).
\end{aligned}
\end{equation}



\subsection{Backward Sampling}
\label{sec: supp-sampling}

As mentioned in section~\ref{sec: sampling}, we propose the backward sampling procedure to generate a collection of trajectories leveraging the distribution information learned by the model. The idea is first to sample the final location $y_{H+T}$ that accounts for most uncertainty in the trajectory. The backward sampling procedure consists of three steps: Evaluation, Sampling, and Completion.

\paragraph{Evaluation} At this stage, we leverage the output $\hat{\pi}$ from the assignment network to determine how we evaluate the distribution of $y_{H+T}$. One can use the component corresponding to $\hat{\pi}_\text{max}$. In our case, we promote multi-modality by computing the distribution as a mixture of top-$6$ components. For handling the latent space $\boldsymbol{v}$, one can apply Monte-Carlo sampling to approximate the integral. In our case, we choose to use the maximum likelihood samples (i.e., $\boldsymbol{v}^{\text{ml}}=\underset{\boldsymbol{v}}{\text{argmax}}p(\boldsymbol{v},z|\boldsymbol{x})$) in equation~\ref{eq: yt-dist} to evaluate the distribution, as shown in Figure~\ref{fig: evaluation}.

\paragraph{Sampling} To allow full exploitation of the distribution information, we first generate a dense grid of candidates that covers the area within $2$ standard deviations of the distribution mean. We adopt a rectangular grid with a resolution of $0.5$ meters for simplicity, as illustrated in Figure~\ref{fig: dense-grid}. One can quickly improve precision by choosing a smaller resolution or clipping the grid area. We then apply the NMS sampling given in Algorithm~\ref{alg: sampling} to sample $M$ candidates from the dense grid considering their circular buffers determined by hyperparameter $r$ and the IoU threshold $\gamma$ (see Figure~\ref{fig: nms-sampling}). Together, the two hyperparameters determine the density of selected candidates. In our practice, we choose $r=1.4$ meters and $\gamma=0\%$. 

\paragraph{Completion} The last step is to complete the intermediate trajectory from the target agent's current position to the sampled final locations. One can easily apply random sampling on each timestep to get the waypoints. Nevertheless, we find trajectories generated by this approach lack auto-consistency and can be non-smooth. To address the problem, we propose a strong assumption that displacement uncertainty is uniform over time. Hence, we can first parameterize an uncertainty distance parameter $u^{(m)}$ for each selected candidate and then use it for computing waypoints for all previous timesteps. Specifically, for sampled candidate $y_{H+T}^{(m)}$ from the distribution $\mathcal{N}\left(\mu_{H+T},\Sigma_{H+T}\right)$, we have
\begin{equation}
    u^{(m)} = L_{H+T}^{-1}\left(y_{H+T}^{(m)}-\mu_{H+T}\right):\Sigma_{H+T}=LL^\intercal,
\end{equation}
where $L$ is the upper triangle Cholesky decomposition of the covariance. For each timestep $t=1,\ldots,T-1$, we have
\begin{equation}
    y_{H+t}^{(m)} = \mu_{H+t} + L_{H+t}\cdot u^{(m)}.
\end{equation}
Finally, we connect intermediate waypoints with the sampled candidates to derive the required trajectory set. Figure~\ref{fig: completion} illustrates the final output from the backward sampling process.

\section{Extensive Qualitative Results}
\label{sec: supp-qual}

\begin{figure*}
    \centering
    \begin{subfigure}{0.45\textwidth}
        \centering
        \includegraphics[width=\textwidth]{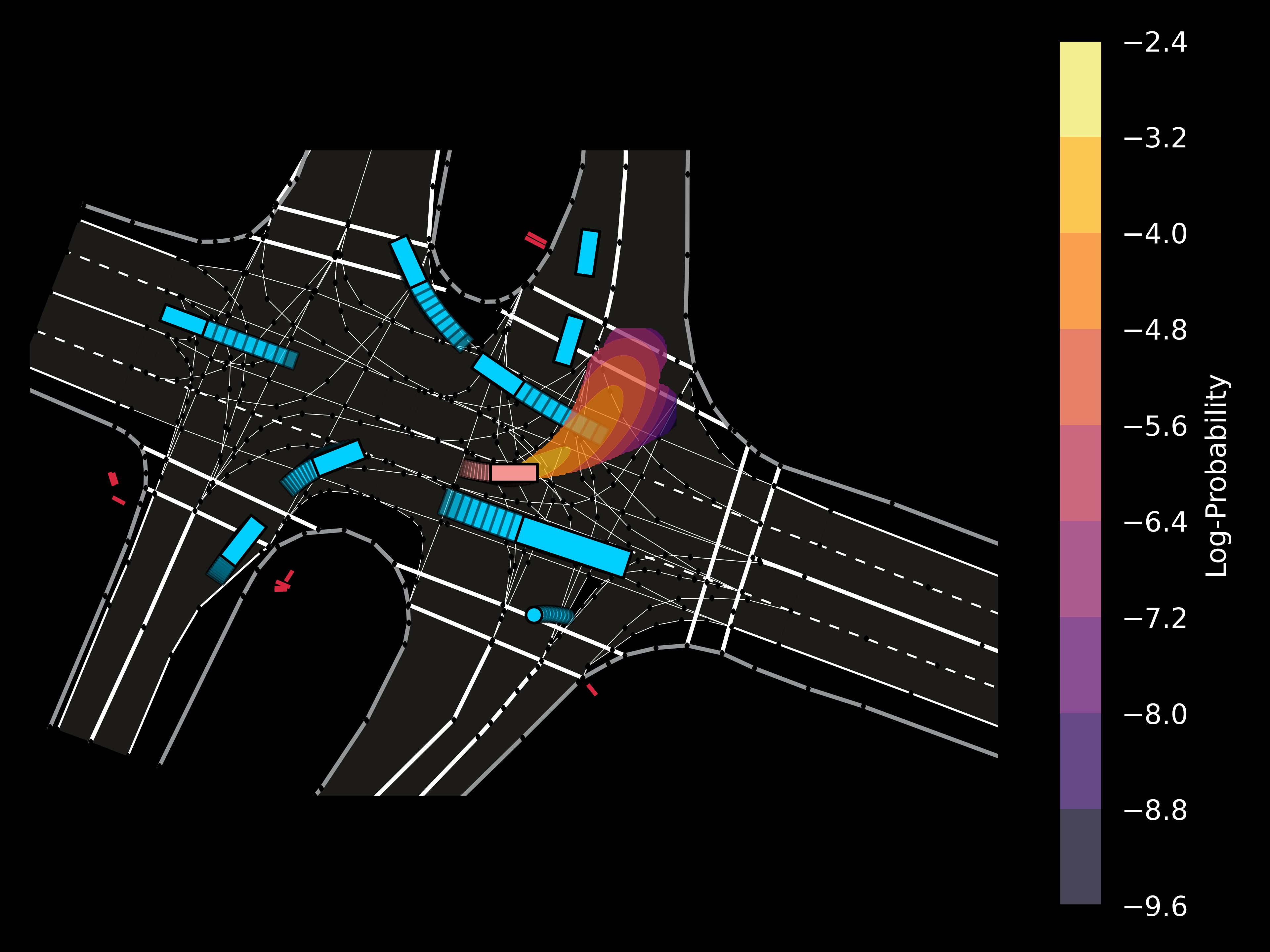}
        \caption{Results on a case from DR\_USA\_Intersection\_GL}
        \label{fig: exp-gl}
    \end{subfigure}
    \hspace{1em}
    \begin{subfigure}{0.45\textwidth}
        \centering
        \includegraphics[width=\textwidth]{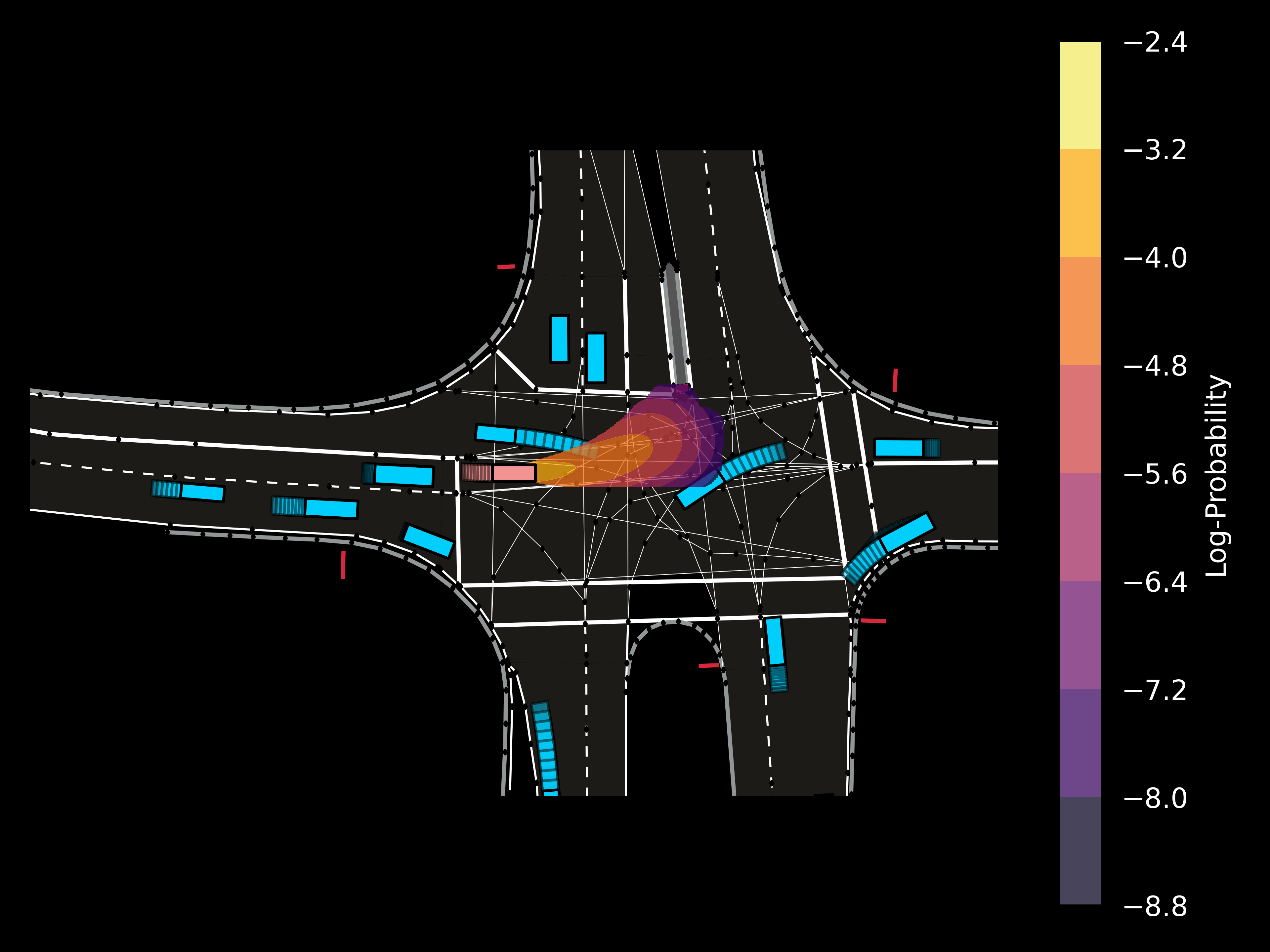}
        \caption{Results on a case from DR\_USA\_Intersection\_MA}
        \label{fig: exp-ma}
    \end{subfigure}
    \caption{\textbf{Representative example visualization of quantified uncertainty on intersections.} The heatmap generated by the SeNeVA model successfully identifies the left-turn intention of drivers in both cases. The predicted distributions conform to the road geometry.}
    \label{fig: exp-intersection}
\end{figure*}

\begin{figure*}
    \centering
    \begin{subfigure}{0.45\textwidth}
        \centering
        \includegraphics[width=\textwidth]{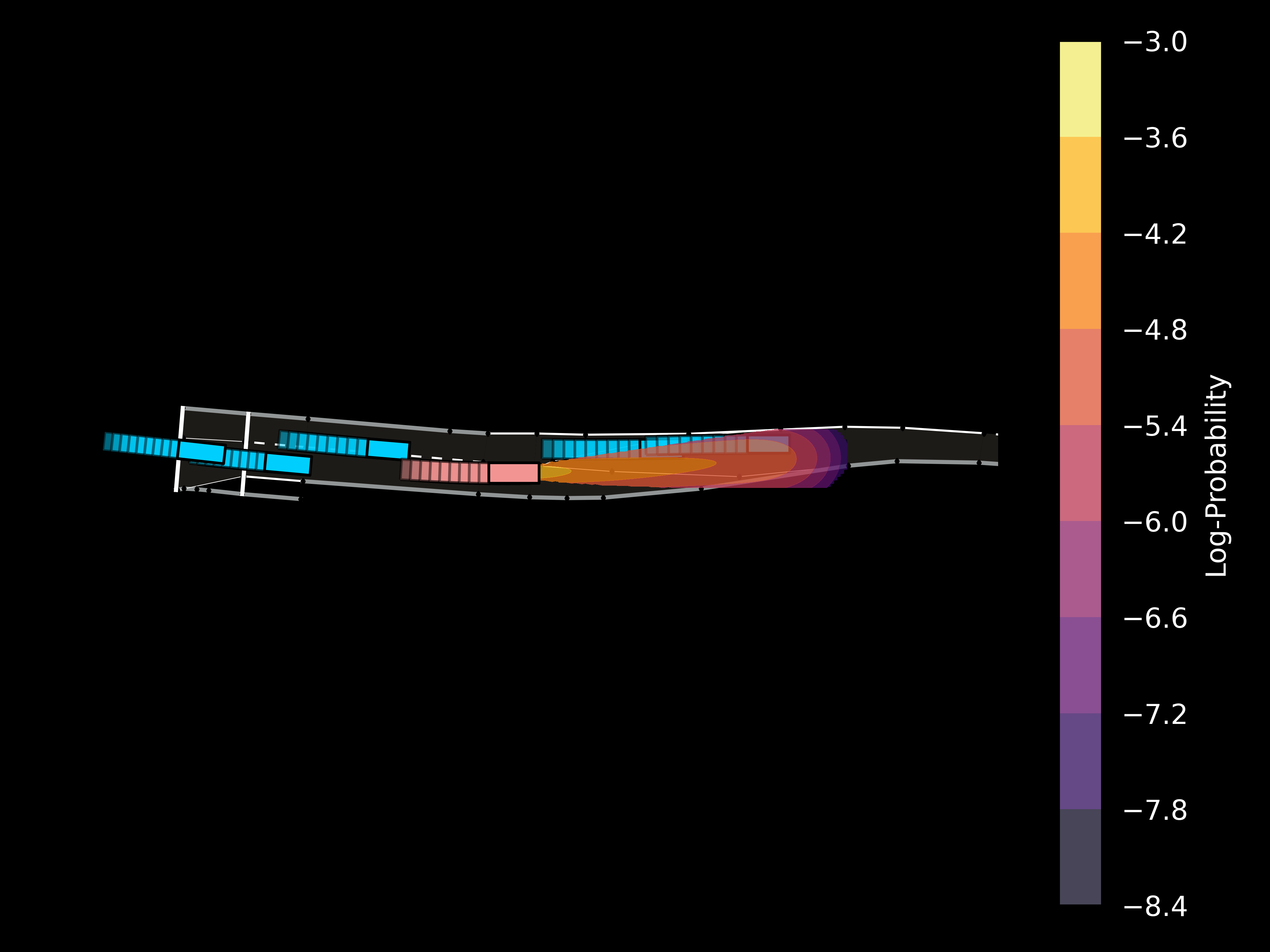}
        \caption{Results on a case from DR\_DEU\_Merging\_MT}
        \label{fig: exp-mt}
    \end{subfigure}
    \hspace{1em}
    \begin{subfigure}{0.45\textwidth}
        \centering
        \includegraphics[width=\textwidth]{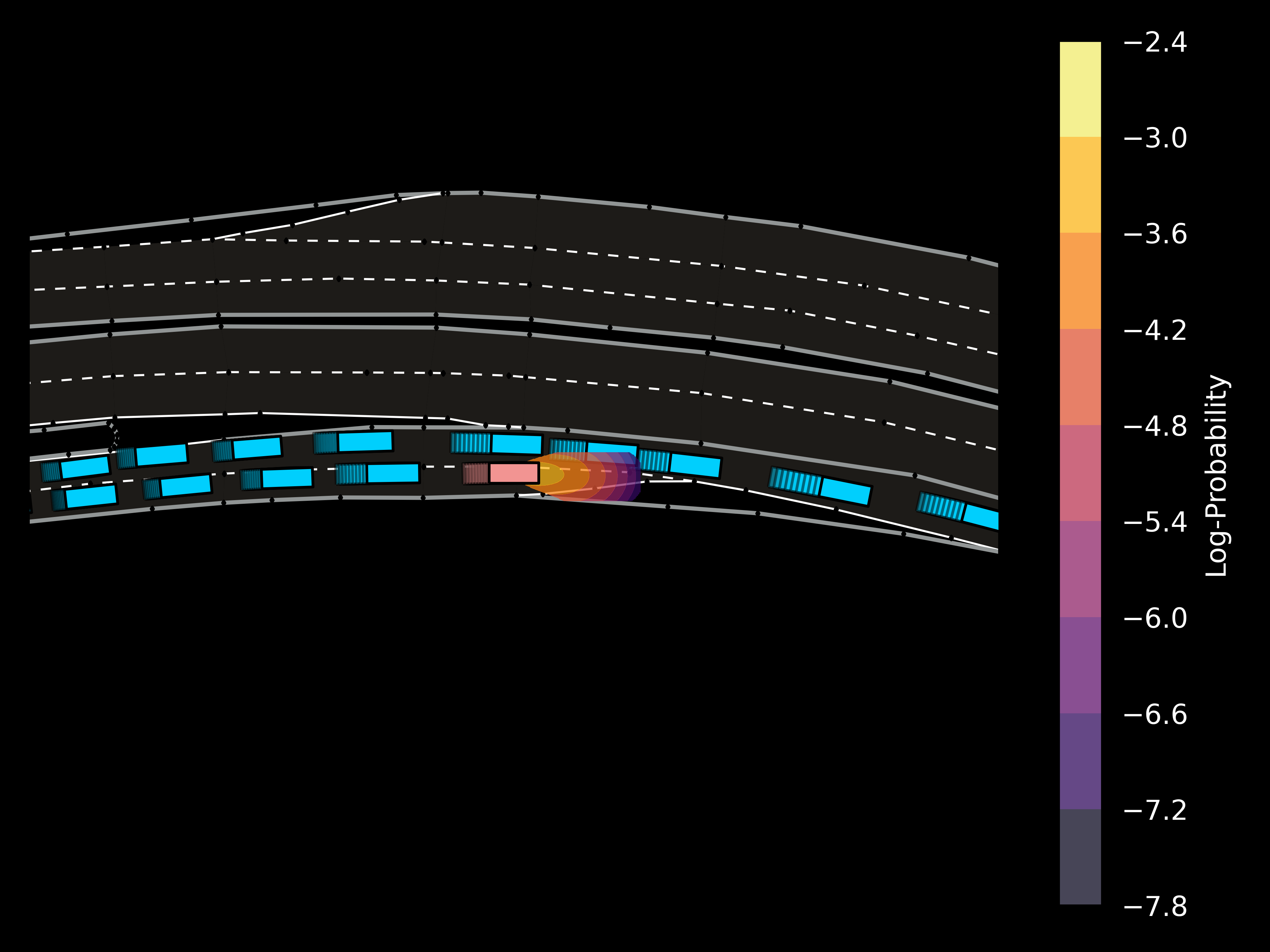}
        \caption{Results on a case from DR\_CHN\_Merging\_ZS0}
        \label{fig: exp-zs}
    \end{subfigure}
    \caption{\textbf{Representative example visualization of quantified uncertainty on intersections.} The model recognizes the existence of surrounding vehicles and predicts with higher certainty that a vehicle will stay hold to avoid collisions.}
    \label{fig: exp-merging}
\end{figure*}

We further visualize the quantified trajectory distributions on some representative cases selected from the INTERACTION dataset. Figure~\ref{fig: exp-intersection} illustrates two examples from unsignalized intersections, where SeNeVA successfully identifies the left-turn intention of the driver and quantifies the distribution of future trajectories that conform to the road geometry. In Figure~\ref{fig: exp-merging}, we visualize two cases in the expressway merging, where the SeNeVA model can anticipate the maneuver of the surrounding vehicles and predict distributions that avoid collisions.

\vfill

\end{document}